\documentclass{article} 
\usepackage{iclr2027_conference,times}

\usepackage{amsmath,amsfonts,bm}

\def\eqref#1{equation~\ref{#1}}

\def\1{\bm{1}}

\DeclareMathAlphabet{\mathsfit}{\encodingdefault}{\sfdefault}{m}{sl}
\SetMathAlphabet{\mathsfit}{bold}{\encodingdefault}{\sfdefault}{bx}{n}

\usepackage{hyperref}       
\usepackage{url}            
\usepackage{booktabs}       
\usepackage{amsfonts}       
\usepackage{nicefrac}       
\usepackage{microtype}      
\usepackage{xcolor}         
\usepackage{colortbl}       
\usepackage{multirow}
\usepackage{graphicx}
\usepackage{float}
\usepackage{capt-of}
\usepackage{amsmath}
\usepackage{tcolorbox}
\usepackage{array}
\usepackage{pgfplots}
\usepackage{wrapfig}
\usepackage{placeins}
\usepackage{zi4}

\pgfplotsset{compat=1.18}

\newcommand{\method}{\textsc{FreeEvolve}}

\title{FreeEvolve: Learning to Evolve Beyond Fixed Loops}

\author{
Lecheng Kong\thanks{Equal contribution.} \\
AWS AI Labs \\
\texttt{jkong@amazon.com}
\And
Like Hui\footnotemark[1] \\
AWS AI Labs \\
\texttt{likehui@amazon.com}
\And
Nikos Kanakaris \\
AWS AI Labs \\
\texttt{nikosk@amazon.com}
\AND
Prithwish Jana\thanks{Work done as an intern at AWS AI Labs.}\\
Georgia Institute of Technology \\
\texttt{pjana7@gatech.edu}
\And
Sahika Genc \\
AWS AI Labs \\
\texttt{sahika@amazon.com}
\And
Narayanan Sadagopan \\
AWS AI Labs \\
\texttt{sdgpn@amazon.com}
}

\iclrfinalcopy 
\begin{document}

\maketitle

\begin{center}
\vspace{-1.5em}
\small\textbf{Project page w/ interactive demo:}
\href{https://freeevolve.github.io/}{{\fontencoding{T1}\fontfamily{zi4}\selectfont
https://freeevolve.github.io/}}
\end{center}


\begin{abstract}
  Agent evolvers automate the design of the prompts, skills and workflows
  around language model agents, yet the optimization process they follow is
  still designed by hand: a fixed search loop decides how candidates are
  evaluated, which are kept and when the search stops. We propose
  \textbf{\method{}}, which automates this process as well. An environment
  specifies the goal, target agent, evaluator, data and resource limits; within
  these limits, the evolver itself decides what to test, how much evidence to
  collect, which candidates to pursue and when to stop. These decisions follow
  an editable \emph{evolution skill}, which we improve through meta-evolution by
  scoring each candidate skill on the fresh target agent it produces. The
  optimization process thus becomes a capability learned from experience rather
  than a loop engineered in advance. On \(\tau^3\)-bench, ARC-AGI-2, ARC-AGI-3
  and Terminal-Bench 2.1, \method{} controls the evolution campaign by itself, yet improves the
  primary held-out metric by \textbf{13.6} points on average and matches or
  exceeds hand-designed evolvers. The learned process
  keeps improving with experience: meta-evolved skills add \(6.9\) points over
  the seed skill on fresh target agents, demonstrating transferability across
  environments.
  \end{abstract}

\section{Introduction}
Large language models (LLMs) have enabled agents to solve a wide range of tasks through reasoning and tool use. However, for a frozen LLM, the performance of an agent depends
  heavily on the harness around it. The harness determines how the model reasons, acts and recovers from failures. Therefore, improving the harness can improve agent performance
  without changing the model weights. Recent studies have started to automate this process by treating prompts, skills and agent workflows as editable components
  \citep{continualHarness2026,lifeHarness2026,skilloptLite2026,harnessCompass2026}. Following this direction, we focus on agent evolvers. An evolver executes a target agent,
  examines its failures and modifies its harness based on the collected evidence.

Although evolvers automate the design of agent harnesses, the optimization
process they follow is still designed by hand. Before evolution begins, a
human-designed search loop fixes which tasks are evaluated, how often
measurements are repeated, how candidates are compared and when the search
stops. On one hand this requires extensive design expertise. On the other hand, a loop encodes assumptions about evaluation noise, task
difficulty, useful curricula and available parallelism. These assumptions vary
across agents and environments and can change as evidence arrives, yet an
evolver that only proposes edits cannot act on that evidence, and adapting the
loop means re-engineering it by hand. We therefore ask: \emph{Can the
optimization process itself be automated, so that the evolver organizes its
own improvement campaign and learns how to do so from experience?}

To this end, we propose \method{}, a framework that automates the
optimization process of agent evolution. The main idea is to treat campaign
organization as a capability that the evolver exercises and learns, rather
than a loop that humans design. Our framework separates the improvement problem from the procedure used to solve it. A bounded
  environment defines the goal, editable target, available data and resource constraints. The evolver is invoked once and remains in control until it returns an improved target.
  After each environment execution, it uses the collected evidence to decide what to do next. This control allows the evolver to adapt the campaign without following an evolution
  loop defined in advance.

 \textbf{Automating the process: campaign autonomy.}
  Campaign autonomy allows the improvement process to adapt as new evidence becomes available. During evolution, the evolver can focus on tasks with frequent failures before
  validating more broadly. It can repeat uncertain comparisons, explore different candidates and reallocate resources based on the observed results. It can also stop when
  additional evidence is unlikely to change the best candidate. These decisions are made as evolution unfolds instead of being fixed in advance by a predefined search loop.

  \textbf{Learning the process: meta-evolution.}
  Campaign autonomy alone does not guarantee an effective improvement process. The evolver still needs guidance on how to use this control. To address this issue, we represent
  the campaign policy as an editable \emph{evolution skill}. We then improve this skill through meta-evolution. Each candidate skill is used to run complete improvement campaigns
  in target-agent environments. The resulting agents and campaign traces are used to evaluate and refine the skill. Meta-evolution therefore produces a self-contained evolution
  skill that can be applied unchanged to new environments without retaining the original agents or execution artifacts.

\begin{figure}[t]
    \centering
    \includegraphics[width=\textwidth]{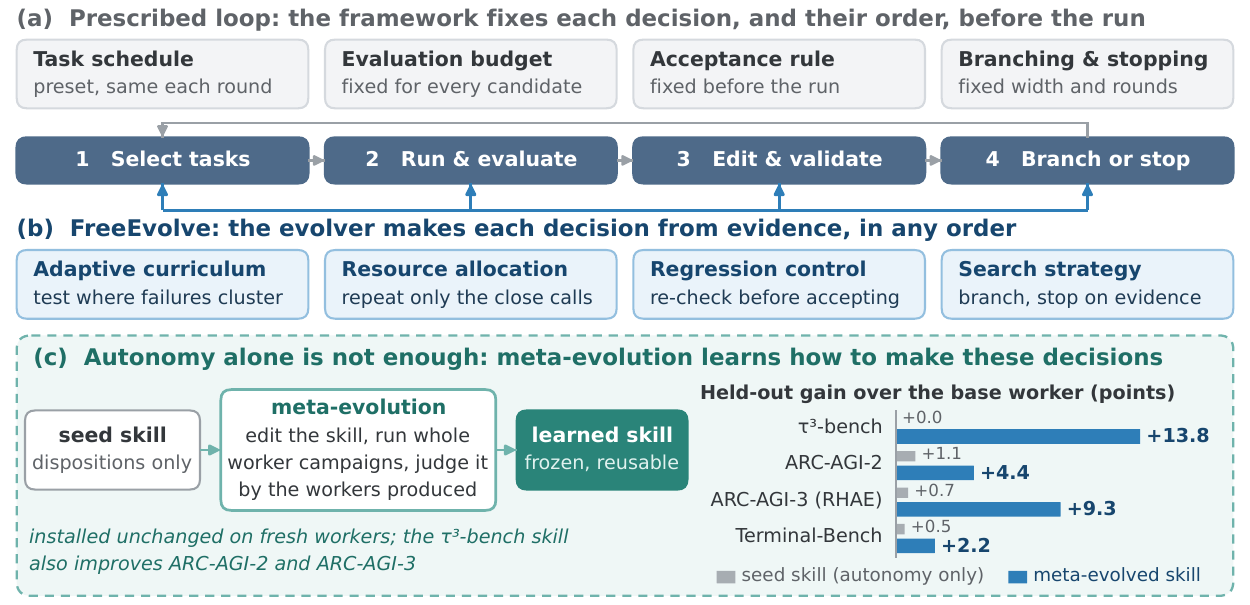}
    \vspace{-6mm}
    \caption{\textbf{From prescribed loops to an automated, learned process.} \textbf{(a) Prescribed loop.} A prescribed loop fixes all campaign decisions and their order before evolution begins. \textbf{(b) Campaign autonomy.} \method{} transfers control of these decisions to the evolver, allowing it to adapt the campaign as new evidence becomes available. \textbf{(c) Meta-evolution.} Campaign autonomy alone provides limited improvements. With the hand-written seed skill, the evolver improves a fresh target agent by at most \(1.1\) points. Meta-evolution evaluates candidate skills through complete improvement campaigns and the target agents they produce. Once frozen and installed on fresh target agents, the learned skills improve held-out performance by \(2.2\) to \(13.8\) points (see Figure~\ref{fig:frozen-evolver}).}
\label{fig:framework}
\vspace{-5mm}
\end{figure}

\textbf{Experimental results.} We evaluate \method{} on \(\tau^3\)-bench \citep{shi2026tau}, ARC-AGI-2
\citep{arcagi22025}, ARC-AGI-3, and Terminal-Bench 2.1
\citep{terminalbench2026}. Without any hand-designed loop, meta-evolution
raises held-out mean performance from \(33.3\%\) to \(51.7\%\), \(55.4\%\) to
\(69.1\%\), \(61.0\%\) to \(81.6\%\), and \(80.9\%\) to \(82.5\%\),
respectively, matching or exceeding hand-designed evolvers on six of seven
metrics. The automated process is also learnable and reusable. Installed on a
fresh worker, the frozen meta-evolved skill outperforms the seed skill on all
four mean metrics and all three available pass@3 metrics, and improves over a
no-data LLM rewrite in all four environments, so execution experience, not
instruction rewriting alone, drives the gain. A skill meta-evolved on \(\tau^3\)-bench also reaches
\(64.3\%\) mean accuracy on ARC-AGI-2 and \(71.0\%\) Relative Human Action
Efficiency (RHAE) on ARC-AGI-3, matching or exceeding the skills meta-evolved
directly on those environments.
\vspace{-2mm}
\paragraph{Contributions.}
\begin{itemize}
\item \textbf{Automating the optimization process.} Conceptually, we
identify the hand-designed search loop as the remaining manual component of
agent evolution and recast it as a campaign policy that the evolver executes
and that can be learned from improvement experience.
\item \textbf{Campaign autonomy and meta-evolution.} Technically, a bounded
environment gives the evolver control of campaign decisions, while a
self-contained evolution skill makes that control improvable through
meta-evolution. Candidate skills are evaluated through the worker campaigns
they organize.
\item \textbf{A learnable and reusable process.} Empirically, without a
hand-designed loop, \method{} matches or exceeds hand-designed evolvers on six
of seven held-out metrics across four environments. Meta-evolution improves
the process itself, which then transfers across environments, yields workers
that transfer to cheaper models, reaches a stronger skill with fewer tokens,
and benefits from more capable evolver models.
\end{itemize}

\section{Related Work}
\textbf{Automating harness design.}
Recent systems automate the design of the prompts, skills, code, and workflows
around a fixed language model
\citep{hu2025adas,zhang2025aflow,continualHarness2026,lifeHarness2026,
livingHarness2026,skilloptLite2026,autoHarness2026,
agenticHarnessEngineering2026}, and evolutionary coding systems improve
programs and agent harnesses through model-generated edits
\citep{sharma2025openevolve,novikov2025alphaevolve,agrawal2025gepa,
metaHarness2026}. They automate \emph{what} is edited; the search process
that decides what to evaluate, which candidates to keep, and when to stop is
typically specified in advance by the system designer.

\textbf{Automating the improver.}
Meta-evolution and self-improvement make the improvement mechanism itself
editable, evolving mutation prompts, program improvers, self-editing agents,
search strategies, context-producing skills, or meta-agents
\citep{promptBreeder2023,stop2024,godelAgent2025,darwinGodelMachine2025,
robeyns2025sica,bilevelAutoresearch2026,metaContextEngineering2026,
liu2026evox,zhang2026hyperagents,aide22026}. 
Meta-Harness, does not do meta-evolution, it lets a coding
proposer read candidate code, scores, and traces before editing
\citep{metaHarness2026}. HyperAgents, like \method{},
evolves at the meta level, jointly evolving task agents and the meta-agents
that improve them, so its procedure can change across outer iterations
\citep{zhang2026hyperagents}. These prior work focus on improving a fixed campaign.

\textbf{Automating the improvement process.}
\method{} extends automation to the process within a run: the evolver
organizes its own campaign as results arrive, and meta-evolution learns the
skill that guides it from completed campaigns. What is learned is neither a
better artifact nor a search procedure for the next run, but a reusable way of
running improvement campaigns. Appendix~\ref{app:related-work} discusses
broader related work.

\section{Method}
\begin{figure*}[t]
  \centering
  \includegraphics[width=\textwidth]{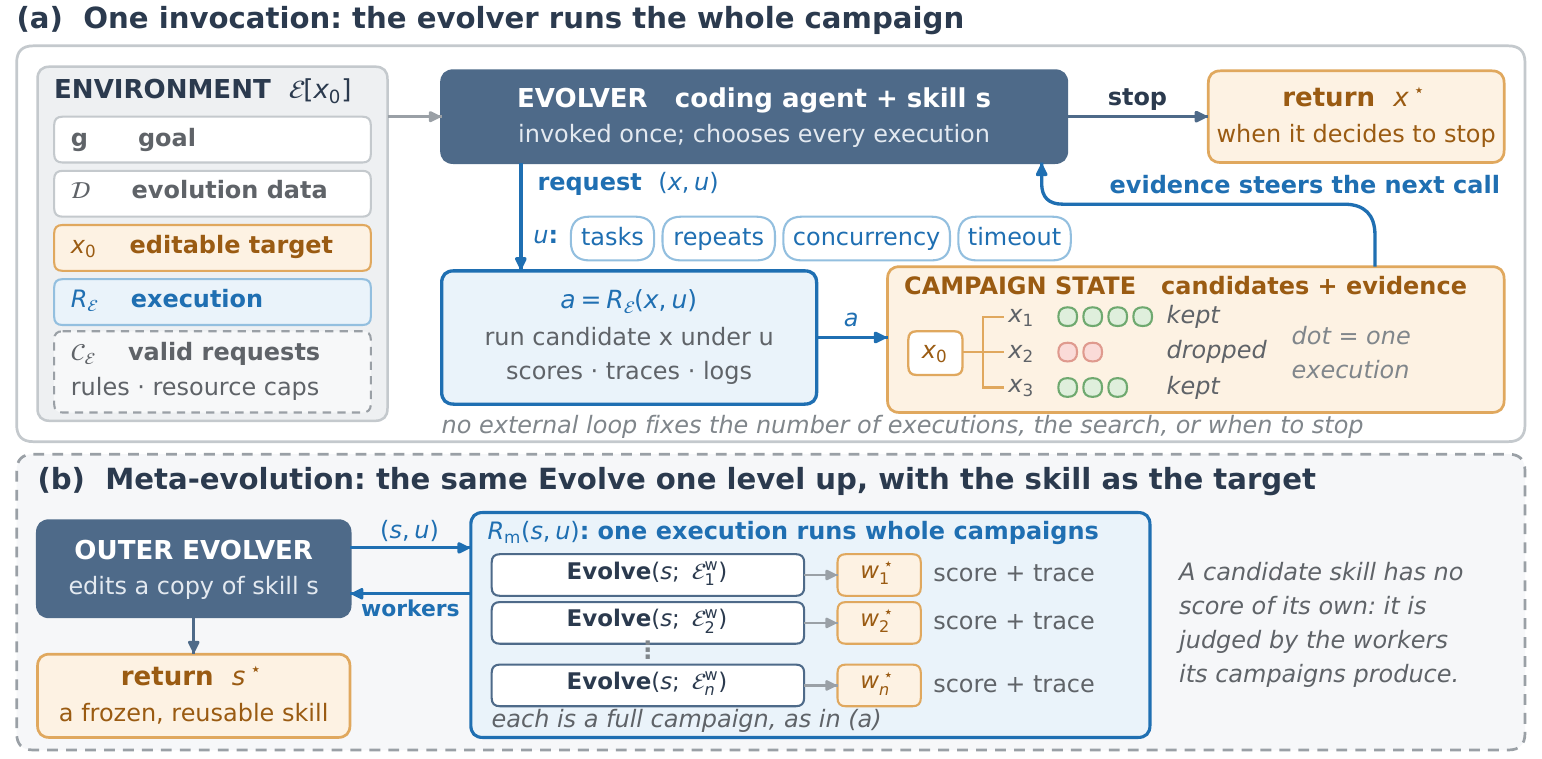}
  \vspace{-6mm}
  \caption{\textbf{The interface at both levels.} (a)~The environment fixes the
  goal, data, editable target, execution function, and valid requests; the
  evolver is invoked once. Each request \((x,u)\) chooses a candidate and how
  to run it (which tasks, how many repeats, what concurrency and timeout). The
  returned artifacts update the campaign state, which steers the next request,
  and the evolver returns \(x^\star\) when it decides to stop. (b)~Meta-evolution
  applies the same \(\operatorname{Evolve}\) with the skill as the target: one
  execution of \(R_{\mathrm{m}}\) runs complete worker campaigns, and a
  candidate skill is judged only by the workers they produce.}
  \label{fig:pipeline}
  \vspace{-5mm}
\end{figure*}

 In this section, we present the proposed \method{} framework. Campaign autonomy changes who organizes the evolution process. In a conventional search loop, the framework
  invokes an evolver or proposer at predefined stages and retains control over evaluation, candidate selection and stopping. In \method{}, the evolver is invoked once and remains
  in control until it returns an improved target. Briefly, the framework operates at two connected levels, as illustrated in Figure~\ref{fig:pipeline}. At the agent level, a
  bounded environment defines the improvement goal, editable target, available data and resource constraints. The evolver follows an evolution skill and organizes the complete
  improvement campaign. It decides what to evaluate, how to allocate resources and when to stop. At the meta level, the evolution skill becomes the editable target. The meta-evolver evaluates candidate skills through complete agent-level campaigns and uses the collected evidence to improve the skill. Both levels use the same environment interface
  and evolution operation.

\subsection{A Bounded Interface for Campaign Autonomy}
\label{sec:interface}

Giving control to the evolver requires a well-defined problem boundary. Within this boundary, the evolver can explore the problem in different ways. However, it cannot change
  the goal, evaluation rules or resource limits. We define an environment initialized with target \(x_0\) as
\begin{equation}
  \mathcal{E}[x_0]
  =
  \left(
    g,
    \mathcal{D},
    x_0,
    R_{\mathcal{E}},
    \mathcal{C}_{\mathcal{E}}
  \right),
  \label{eq:environment}
\end{equation}
where \(g\) is the improvement goal, \(\mathcal{D}\) is the available data
and \(x_0\) is the initial editable target. The execution function
\(R_{\mathcal{E}}\) runs a candidate target \(x\) under an execution request
\(u\) and returns an artifact bundle \(a\):
\begin{equation}
  a = R_{\mathcal{E}}(x,u).
  \label{eq:environment-run}
\end{equation}
The evolver chooses \(u\), and \(\mathcal{C}_{\mathcal{E}}\) defines which
requests are valid. The execution function determines which interactions are
possible; it does not prescribe when or in what order they are used.
\vspace{-1mm}
\paragraph{Worker environments.}
For an agentic benchmark, \(\mathcal{D}\) is the evolution partition of the
benchmark, \(x_0\) is the initial worker harness, and \(g\) is to improve that
harness under the benchmark's evaluation rules. \(R_{\mathcal{E}}\) executes a
candidate harness and returns its scores, trajectories, and logs. The request
\(u\) makes execution choices part of the interface: the evolver may select
tasks, set repetitions and seeds, and allocate concurrency and timeouts, subject
to \(\mathcal{C}_{\mathcal{E}}\), which fixes the benchmark rules and bounds
resources. In a prescribed loop these choices are made before evolution, so the
evolver only ever sees \(R_{\mathcal{E}}(x)\) under a fixed \(u\). Updating prior work to use \(u\) adaptively would require substantial domain knowledge and harness design expertise.

\subsection{The Evolver Controls the Campaign}
\label{sec:evolver}

An evolver is an agent that receives this interface and runs the campaign. We
instantiate it as Claude Code equipped with a
self-contained evolution skill \(s\), a markdown file of instructions. A
complete campaign is one invocation:
\begin{equation}
  x^\star
  =
  \operatorname{Evolve}\!\left(s;\mathcal{E}[x_0]\right)
  \equiv
  s\!\left(
    g,
    \mathcal{D},
    x_0,
    R_{\mathcal{E}},
    \mathcal{C}_{\mathcal{E}}
  \right).
  \label{eq:evolve}
\end{equation}
The framework invokes the evolver once, provides access to
\(R_{\mathcal{E}}\), and takes no further role. The evolver decides which
candidate to edit, which request \((x,u)\) to issue, how the returned artifacts
change its next decision, and when to return \(x^\star\). No external loop fixes
the number of executions, the organization of search, or the stopping
decision. This is our operational definition of campaign autonomy: every
environment execution is chosen by the evolver, and control returns to the
evolver with the resulting artifacts.

Two human-designed components remain, and neither is an evolution loop. Claude
Code's own agent loop (read, act, observe) is generic tool use; it specifies no
evaluation schedule, acceptance rule, or stopping criterion. The seed skill is
written by hand, but it fixes only dispositions: treat the environment as the
source of truth, keep all evidence, continue from the best-supported target,
and use the time budget (Appendix~\ref{app:seed-skill}). Which tasks to
evaluate, how many repetitions to buy, when to branch, and when to accept are
left to the evolver.

\vspace{-1mm}

\paragraph{Worker evolution.}
Let \(\mathcal{E}^{\mathrm{w}}[w_0]\) denote a worker environment initialized
with harness \(w_0\). Worker evolution is an ordinary application of
Equation~\ref{eq:evolve}:
\begin{equation}
  w^\star
  =
  \operatorname{Evolve}
  \left(s;\mathcal{E}^{\mathrm{w}}[w_0]\right).
  \label{eq:worker-evolution}
\end{equation}
Because the evolver initiates every execution, the campaign can follow the
evidence: stable measurements let it move on, close comparisons justify more
repetitions, and failures concentrated in part of the data can be searched
there before a candidate is validated broadly. Whether a given skill organizes
these choices well is a separate question, which we answer by learning the
skill.

\subsection{Meta-Evolution Learns the Campaign Policy}
\label{sec:meta}

Prior meta-evolution makes the search procedure editable across outer
iterations: an outer optimizer produces a new procedure for the next inner run,
whose logic is then fixed while the run executes. \method{} meta-evolves a
different object: an evolution skill that stays active throughout the inner
campaign and decides how to gather evidence and organize search as results
arrive.

Because the skill is itself a text target, meta-evolution needs no new
machinery. We construct a meta-environment whose editable target is an
evolution skill and whose data are worker environments:
\begin{equation}
  \mathcal{D}^{\mathrm{m}}
  =
  \left\{
    \mathcal{E}^{\mathrm{w}}_i[w_i]
  \right\}_{i=1}^{n},
  \label{eq:meta-data}
\end{equation}
\begin{equation}
  \mathcal{E}^{\mathrm{m}}[s_0]
  =
  \left(
    g^{\mathrm{m}},
    \mathcal{D}^{\mathrm{m}},
    s_0,
    R_{\mathrm{m}},
    \mathcal{C}_{\mathrm{m}}
  \right).
  \label{eq:meta-environment}
\end{equation}
A candidate skill is not scored in isolation. It is installed in the evolver
for complete worker-evolution campaigns and judged through the workers and
traces they produce. With \(\mathcal{I}(u)\subseteq\{1,\ldots,n\}\) the worker
environments selected by request \(u\), the meta-level execution function is
\begin{equation}
  R_{\mathrm{m}}(s,u)
  =
  \operatorname{EvaluateWorkers}
  \left(
    \left\{
      \operatorname{Evolve}
      \left(s;\mathcal{E}^{\mathrm{w}}_i[w_i]\right)
      :
      i\in\mathcal{I}(u)
    \right\}
  \right),
  \label{eq:meta-run}
\end{equation}
which returns the resulting workers, their evolution-partition scores, and the
inner campaign records. Meta-evolution is then another application of
Equation~\ref{eq:evolve}:
\begin{equation}
  s^\star
  =
  \operatorname{Evolve}
  \left(
    s_{\mathrm{outer}};
    \mathcal{E}^{\mathrm{m}}[s_0]
  \right).
  \label{eq:meta-evolution}
\end{equation}
The outer evolver edits \(s_0\), decides which inner campaigns to launch, and
uses their outcomes as feedback. \(s_{\mathrm{outer}}\) and \(s_0\) start as
separate copies of the same seed skill, so editing the candidate never changes
the controller. Meta-environments are excluded from \(\mathcal{D}^{\mathrm{m}}\),
which bounds nesting at two levels. Appendix~\ref{app:environment-example}
shows the worker and meta-environments for ARC-AGI-3 side by side.
\vspace{-1mm}
\paragraph{Why the environment--evolver boundary matters.}
  Separating environment assumptions from evolution strategy supports reuse
  across execution settings. On a cluster, inner campaigns finish asynchronously
  across nodes; an evolver tied to a single-machine execution model may leave
  resources idle or require changes to its control logic.

  In \method{}, supporting Slurm requires only an environment-specific
  \(R_{\mathrm{m}}^{\mathrm{slurm}}\), which launches inner campaigns and returns
  their artifacts as they finish, and
  \(\mathcal{C}_{\mathrm{m}}^{\mathrm{slurm}}\), which exposes node allocation
  and concurrency limits. The evolver, evolution skill, and
  \(\operatorname{Evolve}\) operation remain unchanged; the evolver adapts its
  campaign as jobs complete (Appendix~\ref{app:implementation}).

  The same boundary supports skill transfer. Each setting supplies its own
  \(\mathcal{E}[x_0]\), while the skill guides how the evolver uses the available
  capabilities and evidence. A meta-evolved skill can therefore be frozen and
  reused in environments with different tasks, models, or hardware, without
  inheriting source workers or campaign state.

\section{Experiments}
Our central claim is that the policy organizing agent evolution can be learned
rather than fully specified by a human-designed search loop. Campaign autonomy
makes adaptive decisions available, and meta-evolution learns how to use them
from experience. Our experiments answer five questions.
\textbf{Q1:} Can a learned autonomous campaign achieve strong performance
without a prescribed loop? \textbf{Q2:} Does meta-evolution turn campaign
autonomy into a stronger evolver?
\textbf{Q3:} Does the resulting evolution skill generalize beyond its source
environment and model? \textbf{Q4:} How does the evolver use campaign autonomy
as evolution unfolds? \textbf{Q5:} What model capability is required to
exercise campaign autonomy effectively? Appendix Table~\ref{tab:questions}
summarizes the evidence for each.

\subsection{Experimental Setup}

\paragraph{Benchmarks and metrics.}
We evaluate four environments spanning interactive tool use, abstract
reasoning, and terminal control: \(\tau^3\)-bench for conversational agents
over unstructured knowledge \citep{shi2026tau}, ARC-AGI-2 for few-shot grid
transformations \citep{arcagi22025}, ARC-AGI-3 for interactive
abstract-reasoning games \citep{arcagi32026}, and Terminal-Bench 2.1 for
long-horizon command-line tasks \citep{terminalbench2026}. We report mean task
score and pass@3, except on ARC-AGI-3, where we report RHAE. Appendix
Table~\ref{tab:dataset} lists dataset sizes, splits, metrics, and base agents.
All meta-evolution campaigns start from the same task-agnostic seed skill;
Appendix~\ref{app:seed-skill} gives its instructions and the campaign
decisions left open to meta-evolution.
\vspace{-2mm}
\paragraph{Evaluation protocol.}
We measure performance on unseen tasks rather than the best score seen during
search. Each method searches only on the evolution partition, with the same
split and worker configuration within a benchmark, and selects its final
artifact using its own rules before the held-out partition is
revealed. Following the standard search protocol used in GEPA and EvoX
\citep{agrawal2025gepa,liu2026evox}, we perform the search once and
additionally evaluate its resulting worker on the held-out set three times,
reporting the mean and standard deviation (mean \(\pm\) s.d.). Where defined, we also report
pass@3, the fraction of tasks solved at least once across the three runs.

\subsection{Learned Autonomous Campaigns Produce Strong Workers}
\label{sec:main-results}

\definecolor{relativegain}{RGB}{0,92,150}
\begin{table}[!t]
\caption{\textbf{\method{} matches or exceeds
hand-designed evolvers.} Held-out performance (\%) of the worker each method returns; means \(\pm\) s.d.; RHAE: Relative Human
Action Efficiency. EvoX and HyperAgents also meta-evolve their search
procedure. The last row is its
relative gain over the base agent.}
\label{tab:main}
\begin{center}
\small
\setlength{\tabcolsep}{4.5pt}
\begin{tabular}{@{}lccccccc@{}}
\toprule
& \multicolumn{2}{c}{\(\tau^3\)-bench}
& \multicolumn{2}{c}{ARC-AGI-2}
& \multicolumn{1}{c}{ARC-AGI-3}
& \multicolumn{2}{c}{Terminal-Bench 2.1} \\
\cmidrule(lr){2-3}\cmidrule(lr){4-5}\cmidrule(lr){6-6}
\cmidrule(lr){7-8}
Method & Mean & Pass@3 & Mean & Pass@3 & RHAE & Mean & Pass@3 \\
\midrule
Base agent
& 33.3 {\scriptsize\(\pm 5.3\)} & 48.3
& 55.4 {\scriptsize\(\pm 4.0\)} & 75.0
& 61.0 {\scriptsize\(\pm 1.1\)}
& 80.9 {\scriptsize\(\pm 5.0\)} & 88.5 \\
GEPA~\citeyearpar{agrawal2025gepa}
& 35.6 {\scriptsize\(\pm 2.5\)} & 51.7
& 58.0 {\scriptsize\(\pm 5.7\)} & 76.4
& 65.8 {\scriptsize\(\pm 2.6\)}
& \textbf{84.7} {\scriptsize\(\pm 2.4\)} & 91.8 \\
OpenEvolve~\citeyearpar{sharma2025openevolve}
& 35.6 {\scriptsize\(\pm 2.4\)} & 55.2
& 58.6 {\scriptsize\(\pm 5.0\)} & 79.1
& 63.4 {\scriptsize\(\pm 1.5\)}
& 82.5 {\scriptsize\(\pm 1.8\)} & 90.2 \\
EvoX~\citeyearpar{liu2026evox}
& 36.8 {\scriptsize\(\pm 3.5\)} & 58.6
& 64.0 {\scriptsize\(\pm 6.4\)} & 80.0
& 65.1 {\scriptsize\(\pm 3.4\)}
& 82.0 {\scriptsize\(\pm 1.8\)} & 90.2 \\
HyperAgents~\citeyearpar{zhang2026hyperagents}
& 36.8 {\scriptsize\(\pm 5.1\)} & 58.6
& 59.6 {\scriptsize\(\pm 4.2\)} & 77.8
& 72.1 {\scriptsize\(\pm 2.8\)}
& 82.5 {\scriptsize\(\pm 2.1\)} & 90.2 \\
\midrule
\textbf{\method{}}
& \textbf{51.7} {\scriptsize\(\pm 6.0\)} & \textbf{69.0}
& \textbf{69.1} {\scriptsize\(\pm 5.4\)} & \textbf{86.5}
& \textbf{81.6} {\scriptsize\(\pm 3.0\)}
& 82.5 {\scriptsize\(\pm 2.5\)} & \textbf{93.4} \\
{\scriptsize\quad meta-evolution}
& {\scriptsize\color{relativegain}\((+55.3\%)\)}
& {\scriptsize\color{relativegain}\((+40.5\%)\)}
& {\scriptsize\color{relativegain}\((+24.7\%)\)}
& {\scriptsize\color{relativegain}\((+15.3\%)\)}
& {\scriptsize\color{relativegain}\((+33.8\%)\)}
& {\scriptsize\color{relativegain}\((+2.0\%)\)}
& {\scriptsize\color{relativegain}\((+5.5\%)\)} \\
\bottomrule
\end{tabular}
\end{center}
\vspace{-5mm}
\end{table}

To answer \textbf{Q1}, we test whether \method{} can use campaign autonomy to
match the performance of human-designed search loops. We compare the
unmodified worker, GEPA \citep{agrawal2025gepa}, OpenEvolve
\citep{sharma2025openevolve}, EvoX \citep{liu2026evox}, HyperAgents
\citep{zhang2026hyperagents}, and \method{} with meta-evolution. GEPA,
OpenEvolve, and EvoX follow the SkyDiscover settings
\citep{liu2026skydiscover}, and HyperAgents uses its default settings.

Autonomous and prescribed-loop methods allocate evaluations and search effort
differently, making exact cost matching difficult. We therefore use
Table~\ref{tab:main} to compare their improvement frontiers under explicit
stopping rules: each baseline keeps running until improvement stall for 5 iterations, while
\method{} receives a 24-hour meta-evolution budget, the time for different baselines are in Table~\ref{tab:campaign-wall-clock}. The comparison asks how
strong a worker each method can ultimately produce under its own campaign
organization.

Table~\ref{tab:main} shows that a meta-evolved autonomous campaign produces
strong workers across all four environments. \method{} improves the primary
held-out metric by \(13.6\) points on average and outperforms the
expert-designed evolvers on three benchmarks. It also consistently matches or
outperforms the meta-evolution baselines EvoX and HyperAgents, indicating the
value of shifting the target of meta-evolution from a specific search loop to
campaign organization.

Terminal-Bench 2.1 exposes a complementary challenge. \method{} achieves the
best pass@3, while its mean score trails GEPA. Terminal evaluation is
particularly CPU-intensive, so a shared concurrency setting can create
blocking and variable throughput. This contrast reinforces the role of the
environment definition: resource availability and execution behavior should
be exposed better to the evolver.

\subsection{Meta-Evolution Produces a Stronger Evolver}
\label{sec:fresh-worker}

To answer \textbf{Q2}, every skill starts from the same fresh worker and
receives the same resource budget. The \emph{seed} is the initial evolution
skill; the \emph{no-data} variant asks an LLM to improve that skill without
observing execution results; and the \emph{meta-evolved} variant is learned
from completed meta-evolution. No evolved worker or campaign artifact is
carried into these runs.

\begin{figure}[!t]
  \centering
\definecolor{feBase}{HTML}{52514E}%
\definecolor{feSeed}{HTML}{2A78D6}%
\definecolor{feLLM}{HTML}{3A9D78}%
\definecolor{feMeta}{HTML}{E8705F}%
\definecolor{feInk}{HTML}{0B0B0B}%
\definecolor{feMuted}{HTML}{898781}%
\definecolor{feGrid}{HTML}{E1E0D9}%
\begin{tikzpicture}
\pgfplotsset{
  fe panel/.style={
    scale only axis,
    height=4.2cm,
    ymin=-0.6, ymax=3.6,
    ytick={3,2,1,0},
    xtick distance=10,
    xticklabel={\pgfmathprintnumber[assume math mode=true]{\tick}},
    tick label style={font=\scriptsize, text=feInk},
    y tick label style={font=\footnotesize, text=feInk, align=right},
    xmajorgrids,
    major grid style={draw=feGrid, line width=0.4pt},
    axis line style={draw=feMuted!70, line width=0.4pt},
    tick style={draw=feMuted!70},
    major tick length=2pt,
    clip=false,
  },
  fe base/.style={
    only marks, mark=square*, mark size=1.9pt,
    mark options={draw=feBase, fill=white, line width=0.7pt},
  },
  fe seed/.style={
    only marks, mark=triangle*, mark size=2.6pt,
    mark options={draw=white, fill=feSeed, line width=0.4pt},
  },
  fe llm/.style={
    only marks, mark=diamond*, mark size=2.4pt,
    mark options={draw=white, fill=feLLM, line width=0.4pt},
  },
  fe meta/.style={
    only marks, mark=*, mark size=2.3pt,
    mark options={draw=white, fill=feMeta, line width=0.4pt},
  },
  fe whisker/.style={
    error bars/x dir=both,
    error bars/x explicit,
    error bars/error bar style={draw=#1, line width=0.55pt},
    error bars/error mark options={rotate=90, mark size=1.5pt,
      line width=0.55pt, draw=#1},
  },
}
\tikzset{
  fe delta/.style={font=\footnotesize, text=feInk, anchor=center},
  fe head/.style={font=\footnotesize\bfseries, text=feInk, anchor=base,
    yshift=4pt},
  fe sep/.style={draw=feMuted!80, line width=0.5pt, dashed},
  fe val/.style={font=\scriptsize, text=feBase, anchor=west, inner sep=0pt,
    xshift=2pt},
}

\begin{axis}[
  fe panel,
  name=mean,
  width=6.3cm,
  xmin=25, xmax=100,
  yticklabels={
    {\(\tau^3\)-bench},
    {ARC-AGI-2},
    {ARC-AGI-3\\[-1pt]{\scriptsize(RHAE)}},
    {Terminal-\\[-1pt]Bench 2.1}
  },
  legend style={
    at={(axis description cs:0.8,1)},
    anchor=south,
    yshift=15pt,
    legend columns=4,
    draw=none,
    fill=none,
    font=\scriptsize,
    text=feInk,
    /tikz/every even column/.append style={column sep=8pt},
  },
]
\addlegendimage{only marks, mark=square*, mark size=1.9pt,
  mark options={draw=feBase, fill=white, line width=0.7pt}}
\addlegendentry{Base worker}
\addlegendimage{only marks, mark=triangle*, mark size=2.6pt,
  mark options={draw=white, fill=feSeed, line width=0.4pt}}
\addlegendentry{Seed skill}
\addlegendimage{only marks, mark=diamond*, mark size=2.4pt,
  mark options={draw=white, fill=feLLM, line width=0.4pt}}
\addlegendentry{No-data rewrite}
\addlegendimage{only marks, mark=*, mark size=2.3pt,
  mark options={draw=white, fill=feMeta, line width=0.4pt}}
\addlegendentry{Meta-evolved}

\addplot[fe base, fe whisker=feBase, forget plot] coordinates {
  (33.3,3.30) +- (5.3,0)
  (55.4,2.30) +- (4.0,0)
  (61.0,1.30) +- (1.1,0)
  (80.9,0.30) +- (5.0,0)
};
\addplot[fe seed, fe whisker=feSeed, forget plot] coordinates {
  (33.3,3.10) +- (5.3,0)
  (56.5,2.10) +- (1.4,0)
  (61.7,1.10) +- (4.3,0)
  (81.4,0.10) +- (2.5,0)
};
\addplot[fe llm, fe whisker=feLLM, forget plot] coordinates {
  (37.9,2.90) +- (6.0,0)
  (56.8,1.90) +- (3.1,0)
  (64.5,0.90) +- (5.1,0)
  (78.7,-0.10) +- (3.3,0)
};
\addplot[fe meta, fe whisker=feMeta, forget plot] coordinates {
  (47.1,2.70) +- (8.7,0)
  (59.8,1.70) +- (1.9,0)
  (70.3,0.70) +- (3.2,0)
  (83.1,-0.30) +- (4.1,0)
};

\node[fe val] at (axis cs:38.6,3.30) {33.3};
\node[fe val] at (axis cs:55.8,2.70) {47.1};
\node[fe val] at (axis cs:59.4,2.30) {55.4};
\node[fe val] at (axis cs:61.7,1.70) {59.8};
\node[fe val] at (axis cs:62.1,1.30) {61.0};
\node[fe val] at (axis cs:73.5,0.70) {70.3};
\node[fe val] at (axis cs:85.9,0.30) {80.9};
\node[fe val] at (axis cs:87.2,-0.30) {83.1};

\pgfplotsinvokeforeach{0.5,1.5,2.5}{
  \draw[fe sep] (axis cs:25,#1) -- (axis cs:100,#1);
}
\node[fe head] at (axis description cs:0.5,1)
  {Held-out mean score \(\pm\) s.d.\ (\%)};

\node[fe head] at ({axis cs:108,0} |- {axis description cs:0,1}) {\(\Delta\)};
\node[fe delta] at (axis cs:108,3) {+13.8};
\node[fe delta] at (axis cs:108,2) {+3.3};
\node[fe delta] at (axis cs:108,1) {+8.6};
\node[fe delta] at (axis cs:108,0) {+1.7};
\end{axis}

\begin{axis}[
  fe panel,
  name=pass,
  at={(mean.east)},
  anchor=west,
  xshift=1.25cm,
  width=3.6cm,
  xmin=45, xmax=102,
  xtick={50,60,...,100},
  yticklabels={},
]
\addplot[fe base] coordinates {
  (48.3,3.30)
  (75.0,2.30)
  (88.5,0.30)
};
\addplot[fe seed] coordinates {
  (48.3,3.10)
  (74.8,2.10)
  (86.9,0.10)
};
\addplot[fe llm] coordinates {
  (58.6,2.90)
  (74.8,1.90)
  (86.9,-0.10)
};
\addplot[fe meta] coordinates {
  (62.1,2.70)
  (81.1,1.70)
  (90.2,-0.30)
};

\node[fe val, xshift=4pt] at (axis cs:48.3,3.30) {48.3};
\node[fe val, xshift=4pt] at (axis cs:62.1,2.70) {62.1};
\node[fe val, xshift=4pt] at (axis cs:75.0,2.30) {75.0};
\node[fe val, xshift=4pt] at (axis cs:81.1,1.70) {81.1};
\node[fe val, xshift=4pt] at (axis cs:88.5,0.30) {88.5};
\node[fe val, xshift=4pt] at (axis cs:90.2,-0.30) {90.2};

\pgfplotsinvokeforeach{0.5,1.5,2.5}{
  \draw[fe sep] (axis cs:45,#1) -- (axis cs:102,#1);
}
\node[fe head] at (axis description cs:0.5,1) {Held-out pass@3 (\%)};

\node[fe head] at ({axis cs:110,0} |- {axis description cs:0,1}) {\(\Delta\)};
\node[fe delta] at (axis cs:110,3) {+13.8};
\node[fe delta] at (axis cs:110,2) {+6.3};
\node[fe delta, text=feMuted] at (axis cs:110,1) {--};
\node[fe delta] at (axis cs:110,0) {+3.3};
\end{axis}
\end{tikzpicture}
  \vspace{-6mm}
  \caption{\textbf{Meta-evolved skills make stronger evolvers on fresh
workers.} Each skill is frozen and installed on the same fresh base worker
with the same budget. \(\Delta\): meta-evolved minus seed.}
  \label{fig:frozen-evolver}
  \vspace{-6mm}
\end{figure}
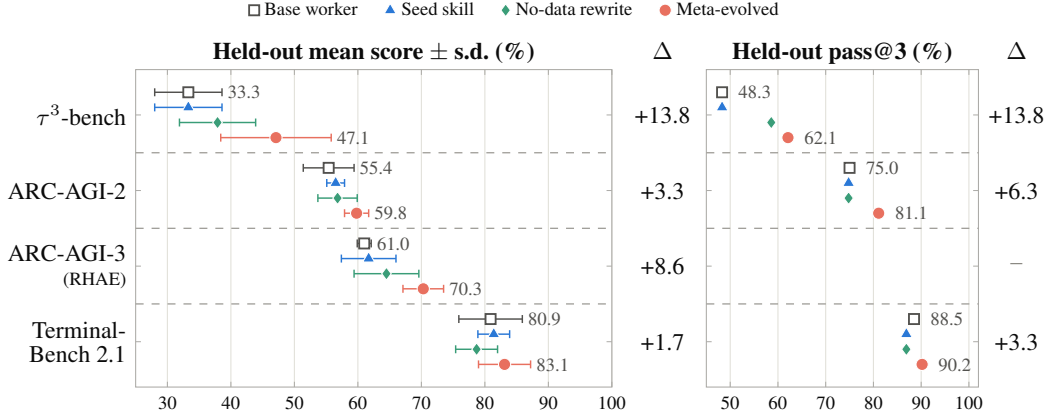

Figure~\ref{fig:frozen-evolver} shows that the seed is already capable of
improving a worker, but only modestly: it gains \(0.6\) points on average
across the four mean metrics. The meta-evolved skill adds a further \(6.9\)
points on average over the seed, improves every available pass@3 metric, and
outperforms the no-data rewrite in every environment. Campaign autonomy makes
a broad set of decisions available; these results show that meta-evolution can
learn how to make those decisions effectively, reducing the need for manual
expert design.

\subsection{Learned Evolution Skills Transfer Across Environments and Models}
\label{sec:cross-environment}

To answer \textbf{Q3}, we test whether an evolution skill, which captures how
to organize improvement rather than a particular worker, remains useful when
installed elsewhere. If so, compute-intensive meta-evolution can be run once
and the resulting evolver reused in new environments.

\definecolor{transfergain}{HTML}{EAF3F9}
\begin{table}[!t]
\caption{\textbf{Learned evolution skills transfer across environments.}
Each row is a skill meta-evolved on the listed environment(s), frozen, and
installed on a fresh worker in each target; shading marks improvement over the
seed. Single-source skills transfer unevenly and lower the Terminal-Bench mean;
meta-evolving on \(\tau^3\)-bench and Terminal-Bench 2.1 (TB2.1) jointly
improves all metrics.}
\label{tab:cross-environment}
\vspace{-2mm}
\begin{center}
\scriptsize
\setlength{\tabcolsep}{2.2pt}
\resizebox{\columnwidth}{!}{%
\begin{tabular}{@{}lccccccc@{}}
\toprule
& \multicolumn{2}{c}{Target: \(\tau^3\)-bench}
& \multicolumn{2}{c}{Target: ARC-AGI-2}
& \multicolumn{1}{c}{Target: ARC-AGI-3}
& \multicolumn{2}{c}{Target: TB 2.1} \\
\cmidrule(lr){2-3}\cmidrule(lr){4-5}\cmidrule(lr){6-6}
\cmidrule(lr){7-8}
Meta-evolution environment(s)
& Mean & Pass@3 & Mean & Pass@3 & RHAE & Mean & Pass@3 \\
\midrule
Minimal seed
& 33.3 {\scriptsize\(\pm 5.3\)} & 48.3
& 56.5 {\scriptsize\(\pm 1.4\)} & 74.8
& 61.7 {\scriptsize\(\pm 4.3\)}
& 81.4 {\scriptsize\(\pm 2.5\)} & 86.9 \\
\midrule
\(\tau^3\)-bench
& \cellcolor{transfergain}\textbf{47.1 {\scriptsize\(\pm 8.7\)}}
& \cellcolor{transfergain}\textbf{62.1}
& \cellcolor{transfergain}\textbf{64.3 {\scriptsize\(\pm 2.8\)}}
& \cellcolor{transfergain}\textbf{83.8}
& \cellcolor{transfergain}\textbf{71.0 {\scriptsize\(\pm 6.4\)}}
& 76.5 {\scriptsize\(\pm 7.4\)}
& \cellcolor{transfergain}90.2 \\
ARC-AGI-2
& \cellcolor{transfergain}44.8 {\scriptsize\(\pm 3.4\)}
& \cellcolor{transfergain}55.2
& \cellcolor{transfergain}59.8 {\scriptsize\(\pm 1.9\)}
& \cellcolor{transfergain}81.1
& \cellcolor{transfergain}62.5 {\scriptsize\(\pm 3.6\)}
& 74.3 {\scriptsize\(\pm 5.0\)}
& 85.2 \\
\(\tau^3\)-bench + TB2.1
& \cellcolor{transfergain}46.0 {\scriptsize\(\pm 8.7\)}
& \cellcolor{transfergain}51.7
& \cellcolor{transfergain}58.0 {\scriptsize\(\pm 1.9\)}
& \cellcolor{transfergain}81.1
& \cellcolor{transfergain}66.9 {\scriptsize\(\pm 4.8\)}
& \cellcolor{transfergain}\textbf{82.5 {\scriptsize\(\pm 4.1\)}}
& \cellcolor{transfergain}\textbf{91.8} \\
\bottomrule
\end{tabular}
}
\end{center}
\vspace{-5mm}
\end{table}

We freeze each skill learned in a separate meta-evolution campaign and
install it in a fresh target environment (Table~\ref{tab:cross-environment}).
Transfer is positive on most target metrics, and some transferred skills
approach or exceed the in-environment result. Both single-source skills,
however, reduce the Terminal-Bench 2.1 mean, and the ARC-AGI-2 skill also
reduces its pass@3, suggesting a transfer boundary when the target
environment, including its hardware and execution behavior, changes
substantially. Joint meta-evolution on \(\tau^3\)-bench and Terminal-Bench
2.1 yields a skill that improves every reported target metric over the seed,
so exposing meta-evolution to more diverse environments can produce a more
robust evolver.

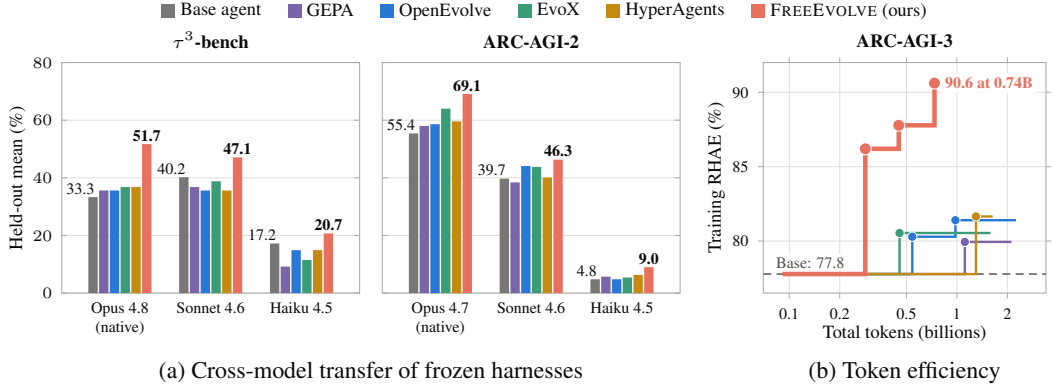
\begin{figure}[!t]
  \centering
\definecolor{tfInk}{HTML}{0B0B0B}
\definecolor{tfMuted}{HTML}{898781}
\definecolor{tfGrid}{HTML}{E1E0D9}
\definecolor{tfBase}{HTML}{52514E}
\definecolor{tfOpen}{HTML}{2A78D6}
\definecolor{tfGepa}{HTML}{7A65A8}
\definecolor{tfEvoX}{HTML}{3A9D78}
\definecolor{tfHyper}{HTML}{C58A12}
\definecolor{tfFree}{HTML}{E8705F}
\colorlet{tfBaseTint}{tfBase!78}

\begin{tikzpicture}
\pgfplotsset{
  tf panel/.style={
    scale only axis,
    height=3.05cm,
    axis line style={draw=tfMuted!70, line width=0.4pt},
    tick style={draw=tfMuted!70},
    major tick length=2pt,
    major grid style={draw=tfGrid, line width=0.4pt},
    tick label style={font=\tiny, text=tfInk},
    label style={font=\scriptsize, text=tfInk},
    ylabel shift=-4pt,
    title style={font=\scriptsize\bfseries, text=tfInk, yshift=-4pt},
    clip=false,
  },
  tf bars/.style={
    tf panel,
    width=3.94cm,
    ymin=0, ymax=80,
    ytick={0,20,40,60,80},
    symbolic x coords={Native,Sonnet,Haiku},
    xtick=data,
    enlarge x limits=0.32,
    ymajorgrids,
    x tick style={draw=none},
    x tick label style={align=center},
  },
  tf bar/.style 2 args={
    ybar, bar width=3.53pt, bar shift=#1, fill=#2, draw=none,
  },
  tf swatch/.style={
    legend image code/.code={
      \fill[##1, draw=none] (0cm,-0.09cm) rectangle (0.18cm,0.09cm);
    },
  },
  tf line/.style={const plot, mark=none, line width=0.9pt, forget plot},
}
\tikzset{
  tf base label/.style={font=\tiny, text=tfInk, anchor=south east,
    inner sep=1pt, xshift=-8.35pt},
  tf free label/.style={font=\tiny\bfseries, text=tfInk, anchor=south,
    inner sep=1pt, xshift=10.11pt},
  tf dot/.style={circle, fill=#1, draw=white, line width=0.35pt,
    minimum size=3.5pt, inner sep=0pt},
  tf sub/.style={font=\footnotesize, text=tfInk, anchor=north, inner sep=1pt},
}

\begin{axis}[
  tf bars,
  name=tau,
  title={\(\tau^3\)-bench},
  ylabel={Held-out mean (\%)},
  xticklabels={{Opus 4.8\\(native)},{Sonnet 4.6},{Haiku 4.5}},
  legend style={
    at={(axis description cs:1.628,1)},
    anchor=south,
    yshift=12pt,
    legend columns=6,
    draw=none,
    fill=none,
    font=\scriptsize,
    text=tfInk,
    /tikz/every even column/.append style={column sep=8pt},
  },
  legend cell align=left,
]
\addlegendimage{tf swatch, fill=tfBaseTint}
\addlegendentry{Base agent}
\addlegendimage{tf swatch, fill=tfGepa}
\addlegendentry{GEPA}
\addlegendimage{tf swatch, fill=tfOpen}
\addlegendentry{OpenEvolve}
\addlegendimage{tf swatch, fill=tfEvoX}
\addlegendentry{EvoX}
\addlegendimage{tf swatch, fill=tfHyper}
\addlegendentry{HyperAgents}
\addlegendimage{tf swatch, fill=tfFree}
\addlegendentry{\method{} (ours)}

\addplot[tf bar={-10.11pt}{tfBaseTint}, forget plot] coordinates {
  (Native,33.3) (Sonnet,40.2) (Haiku,17.2)};
\addplot[tf bar={-6.07pt}{tfGepa}, forget plot] coordinates {
  (Native,35.6) (Sonnet,36.8) (Haiku,9.2)};
\addplot[tf bar={-2.02pt}{tfOpen}, forget plot] coordinates {
  (Native,35.6) (Sonnet,35.6) (Haiku,14.9)};
\addplot[tf bar={2.02pt}{tfEvoX}, forget plot] coordinates {
  (Native,36.8) (Sonnet,38.8) (Haiku,11.5)};
\addplot[tf bar={6.07pt}{tfHyper}, forget plot] coordinates {
  (Native,36.8) (Sonnet,35.6) (Haiku,14.9)};
\addplot[tf bar={10.11pt}{tfFree}, forget plot] coordinates {
  (Native,51.7) (Sonnet,47.1) (Haiku,20.7)};

\node[tf base label] at (axis cs:Native,33.3) {33.3};
\node[tf base label] at (axis cs:Sonnet,40.2) {40.2};
\node[tf base label] at (axis cs:Haiku,17.2) {17.2};
\node[tf free label] at (axis cs:Native,51.7) {51.7};
\node[tf free label] at (axis cs:Sonnet,47.1) {47.1};
\node[tf free label] at (axis cs:Haiku,20.7) {20.7};
\end{axis}

\begin{axis}[
  tf bars,
  name=arc,
  at={(tau.south east)},
  anchor=south west,
  xshift=0.3cm,
  title={ARC-AGI-2},
  yticklabels={},
  xticklabels={{Opus 4.7\\(native)},{Sonnet 4.6},{Haiku 4.5}},
]
\addplot[tf bar={-10.11pt}{tfBaseTint}, forget plot] coordinates {
  (Native,55.4) (Sonnet,39.7) (Haiku,4.8)};
\addplot[tf bar={-6.07pt}{tfGepa}, forget plot] coordinates {
  (Native,58.0) (Sonnet,38.4) (Haiku,5.7)};
\addplot[tf bar={-2.02pt}{tfOpen}, forget plot] coordinates {
  (Native,58.6) (Sonnet,44.1) (Haiku,4.8)};
\addplot[tf bar={2.02pt}{tfEvoX}, forget plot] coordinates {
  (Native,64.0) (Sonnet,43.8) (Haiku,5.4)};
\addplot[tf bar={6.07pt}{tfHyper}, forget plot] coordinates {
  (Native,59.6) (Sonnet,40.1) (Haiku,6.3)};
\addplot[tf bar={10.11pt}{tfFree}, forget plot] coordinates {
  (Native,69.1) (Sonnet,46.3) (Haiku,9.0)};

\node[tf base label] at (axis cs:Native,55.4) {55.4};
\node[tf base label] at (axis cs:Sonnet,39.7) {39.7};
\node[tf base label] at (axis cs:Haiku,4.8) {4.8};
\node[tf free label] at (axis cs:Native,69.1) {69.1};
\node[tf free label] at (axis cs:Sonnet,46.3) {46.3};
\node[tf free label] at (axis cs:Haiku,9.0) {9.0};
\end{axis}

\begin{axis}[
  tf panel,
  name=cost,
  at={(arc.south east)},
  anchor=south west,
  xshift=1.10cm,
  width=3.74cm,
  title={ARC-AGI-3},
  xmode=log,
  log basis x=10,
  xmin=7e7, xmax=3.4e9,
  ymin=0.765, ymax=0.92,
  xtick={1e8,2e8,5e8,1e9,2e9},
  xticklabels={0.1,0.2,0.5,1,2},
  ytick={0.80,0.85,0.90},
  yticklabels={80,85,90},
  xlabel={Total tokens (billions)},
  ylabel={Training RHAE (\%)},
  xlabel style={yshift=4.5pt},
  grid=major,
  tick align=outside,
]
\addplot[tfBaseTint, densely dashed, line width=0.7pt, forget plot,
  domain=7e7:3.4e9] {0.7778};
\node[font=\tiny, text=tfBase, anchor=south west, fill=white, inner sep=1pt]
  at (axis cs:8.0e7,0.7795) {Base: 77.8};

\addplot[tf line, tfOpen] coordinates {
  (299997630,0.7778) (541924236,0.8029) (765040411,0.8029)
  (979442892,0.8140) (1230211815,0.8140) (1436003196,0.8140)
  (1685398761,0.8140) (1966493286,0.8140) (2249562285,0.8140)};
\addplot[tf line, tfGepa] coordinates {
  (244953667,0.7778) (245198513,0.7778) (567737322,0.7778)
  (852128413,0.7778) (1115227089,0.7994) (1374219406,0.7994)
  (2116202656,0.7994)};
\addplot[tf line, tfEvoX] coordinates {
  (293853334,0.7778) (454878196,0.8054) (646284776,0.8054)
  (842088384,0.8054) (1107245078,0.8054) (1264125752,0.8054)
  (1425483310,0.8054) (1588785292,0.8054)};
\addplot[tf line, tfHyper] coordinates {
  (281868245,0.7778) (592964714,0.7778) (943401194,0.7778)
  (1299384568,0.8165) (1635293736,0.8165)};
\addplot[tf line, tfFree, line width=1.7pt] coordinates {
  (91055370,0.7778) (284245847,0.8620) (450150516,0.8778)
  (597082615,0.8778) (735099650,0.9062)};

\node[tf dot=tfOpen] at (axis cs:541924236,0.8029) {};
\node[tf dot=tfOpen] at (axis cs:979442892,0.8140) {};
\node[tf dot=tfGepa] at (axis cs:1115227089,0.7994) {};
\node[tf dot=tfEvoX] at (axis cs:454878196,0.8054) {};
\node[tf dot=tfHyper] at (axis cs:1299384568,0.8165) {};
\node[tf dot=tfFree, minimum size=4.5pt] at (axis cs:284245847,0.8620) {};
\node[tf dot=tfFree, minimum size=4.5pt] at (axis cs:450150516,0.8778) {};
\node[tf dot=tfFree, minimum size=4.5pt] at (axis cs:735099650,0.9062) {};

\node[font=\tiny\bfseries, text=tfFree, anchor=west, fill=white,
  fill opacity=0.92, text opacity=1, inner sep=1.5pt]
  at (axis cs:8.0e8,0.9062) {90.6 at 0.74B};
\end{axis}

\node[tf sub, xshift=4.090cm, yshift=-0.86cm] at (tau.south west)
  {(a) Cross-model transfer of frozen harnesses};
\node[tf sub, yshift=-0.86cm] at (cost.south)
  {(b) Token efficiency};
\end{tikzpicture}
  \vspace{-8mm}
  \caption{\textbf{Evolved harnesses transfer to cheaper worker models, and
\method{} learns a strong and efficient skill.} (a)~Each method evolves
the harness with the native model; the frozen harness then runs unchanged with Sonnet 4.6 and Haiku
4.5. \method{} is the strongest harness in all six settings and keeps a
\(5.3\)-point average gain over the base agent on the four non-native ones.
(b)~Best-so-far ARC-AGI-3 training RHAE of the worker each method's skill
produces, versus the total tokens used to produce that skill.}
  \label{fig:cross-model-transfer}
  \label{fig:arc3-cost-curve}
  \vspace{-5mm}
\end{figure}

We also test whether evolved worker harnesses capture task improvements
rather than model-specific behavior: each method's harness, evolved with the
native Opus worker, runs unchanged with Sonnet 4.6 and Haiku 4.5
(Figure~\ref{fig:cross-model-transfer}a). Across the four non-native settings,
\method{} keeps an average gain of \(5.3\) points over the base harness and
remains the strongest evolved harness, whereas several baseline gains on
\(\tau^3\)-bench shrink or reverse. \method{} workers can therefore be evolved
with a stronger model and deployed with a cheaper one without re-evolution.

Figure~\ref{fig:arc3-cost-curve}b measures how effectively each method converts
its total token budget, including all evolver, evaluation, and other model
calls, into a stronger evolver skill, scored by the training RHAE of the
worker the skill produces. By \(0.74\)B tokens, \method{} has produced a skill
whose worker reaches a score that no baseline attains with more than twice the
total token cost.

\subsection{Model Capability Supports Effective Campaign Control}
\label{sec:evolver-model-scaling}

\begin{wraptable}{R}{0.39\textwidth}
  \vspace{-4mm}
  \setlength{\abovecaptionskip}{0pt}
  \setlength{\belowcaptionskip}{3pt}
  \centering
  \caption{Evolver-model ablation on \mbox{\(\tau^3\)-bench}.}
  \label{tab:evolver-model-scaling-main}
  \footnotesize
  \setlength{\tabcolsep}{4pt}
  \renewcommand{\arraystretch}{1.08}
  \begin{tabular}{@{}lcc@{}}
    \toprule
    \textbf{Evolver model}
    & \textbf{Mean (\%)}
    & \textbf{Pass@3} \\
    \midrule
    No meta-evolution & 33.3 {\scriptsize\(\pm 5.3\)} & 48.3 \\
    \midrule
    Haiku 4.5         & 32.2 {\scriptsize\(\pm 8.7\)} & 48.3 \\
    Sonnet 4.6        & 47.1 {\scriptsize\(\pm 4.0\)} & 62.1 \\
    Opus 4.6          & 40.2 {\scriptsize\(\pm 2.0\)} & 55.2 \\
    Opus 4.7          & 49.4 {\scriptsize\(\pm 8.0\)} & \textbf{69.0} \\
    Opus 4.8          & \textbf{51.7 {\scriptsize\(\pm 6.0\)}} & \textbf{69.0} \\
    \bottomrule
  \end{tabular}
  \vspace{-4mm}
\end{wraptable}

\textbf{Q5} asks what model capability is needed to turn campaign autonomy
into held-out gains. We hold the \(\tau^3\)-bench worker and meta-evolution
setup fixed, vary only the model executing the evolver from Haiku 4.5 to
Opus 4.8, and compare the resulting workers only on held-out mean and pass@3.

Table~\ref{tab:evolver-model-scaling-main} reveals a clear transition. Haiku
4.5 does not improve either held-out metric; beginning with Sonnet 4.6,
meta-evolution consistently improves held-out performance, with the strongest
Opus models producing the best workers. Held-out improvement generally
increases with newer and stronger models, although Sonnet 4.6 notably
outperforms Opus 4.6. This result also identifies a limitation:
producing held-out gains with \method{} requires sufficient model capability. Sonnet 4.6 already demonstrates this capability; we leave a broader
study of this boundary to future work.

\FloatBarrier
\subsection{Autonomous Campaigns Adapt Evaluation and Search}
\label{sec:skill-analysis}

\begin{wraptable}{R}{0.42\textwidth}
  \vspace{-2mm}
  \setlength{\abovecaptionskip}{0pt}
  \setlength{\belowcaptionskip}{3pt}
  \centering
  \caption{Fixed-decision ablations on \mbox{\(\tau^3\)-bench}.}
  \label{tab:fixed-u-main}
  \footnotesize
  \setlength{\tabcolsep}{4pt}
  \renewcommand{\arraystretch}{1.05}
  \begin{tabular}{@{}lcc@{}}
    \toprule
    \textbf{Campaign control}
    & \textbf{Mean (\%)}
    & \textbf{Tokens (B)} \\
    \midrule
    Controlled \(u\)    & \textbf{47.1} & \textbf{1.19} \\
    Full set            & 39.1          & 0.89 \\
    Concurrency \(=24\) & 43.7          & 1.78 \\
    No timeout          & 44.8          & 1.53 \\
    \bottomrule
  \end{tabular}
  \vspace{-4mm}
\end{wraptable}

To answer \textbf{Q4}, we hold the learned skill and resource budget fixed
while removing its control over one environment-facing decision at a time. Table~\ref{tab:fixed-u-main} examines how the learned skill uses its
campaign control by fixing one decision at a time. Always evaluating the full
set makes the evolver wait for expensive probes, reducing the number of
decisions it can make within the time budget. Fixed concurrency and the
absence of timeouts leave resources occupied by slow or failed executions,
using more total tokens but producing a weaker worker. The benefit is thus not
one universally optimal configuration; it is the ability to reorganize
evaluation as the campaign unfolds.

\paragraph{Qualitative campaign analysis.}

The fixed-decision interventions isolate evaluation choices, but autonomy also
governs the organization of search. We therefore trace one ARC-AGI-2
worker-evolution campaign using a frozen meta-evolved skill.
\begin{figure}[!t]
  \centering
  \includegraphics[width=\textwidth]{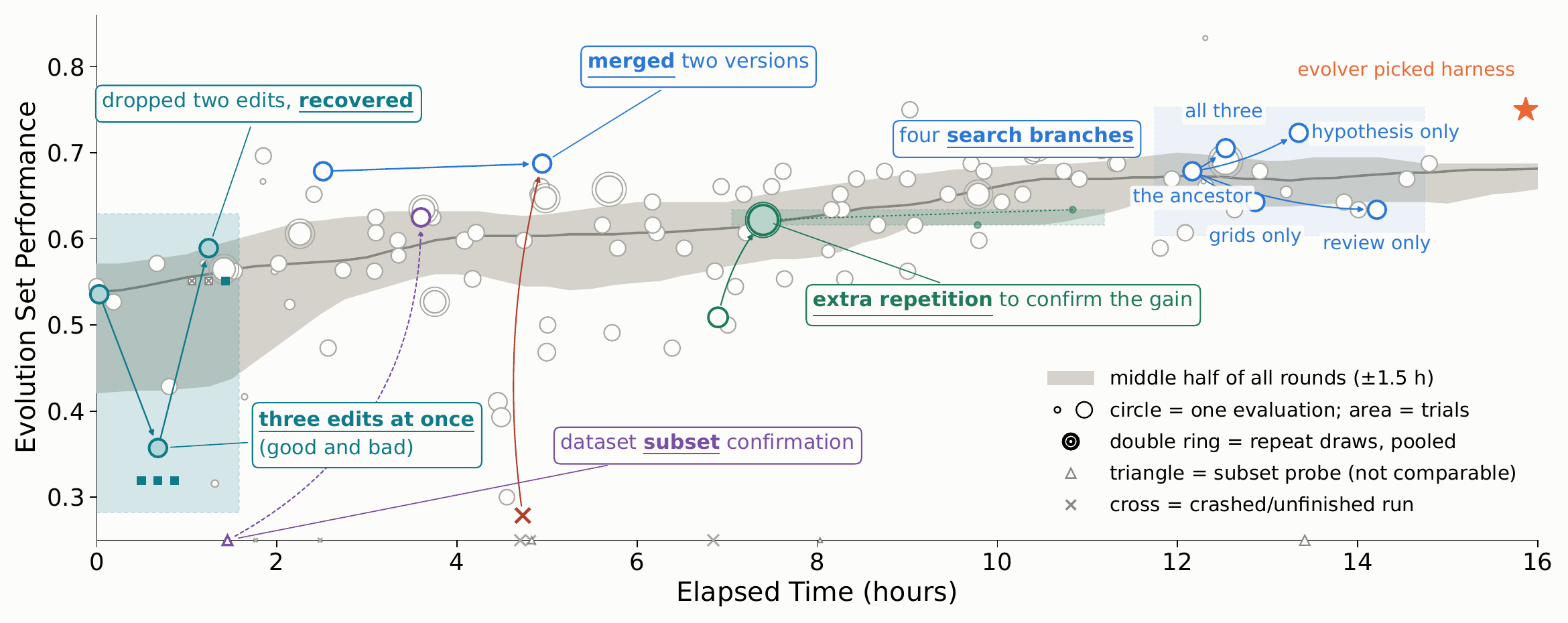}
  \vspace{-7mm}
  \caption{\textbf{The evolver reorganizes evaluation and search as evidence
  arrives.} One \mbox{ARC-AGI-2} worker-evolution campaign under a frozen
  meta-evolved skill: the evolution-set score of every evaluation against
  elapsed hours (markers as in the legend). Annotations mark five decisions
  the evolver made on its own: undoing a three-edit regression while keeping
  the useful helper library, probing a 2\% subset to show that a truncation
  came from the agent's own prompt rather than the platform, merging the best
  code with a fix from a weaker run, repeating a promising version to confirm
  its gain, and branching an ancestor into four variants. The star marks the
  harness it selects.}
  \label{fig:arc2-campaign-behaviors}
  \vspace{-5mm}
\end{figure}
Figure~\ref{fig:arc2-campaign-behaviors} shows the evolver exercising this
control online: it changes evaluation resolution, opens distinct branches,
combines discoveries, and selects the final worker as evidence accumulates.
These behaviors are not hard-coded stages of the framework; the evolver
invokes each behavior when the campaign requires it.

To identify what meta-evolution contributes to these decisions, we compare
the four meta-evolved skills with their common seed. The seed provides broad
dispositions---use the available time, preserve evidence, and continue from
the best worker---while leaving the organization of the campaign open. The
evolved skills turn those dispositions into operational policies
(Appendix~\ref{app:additional-results},
Figure~\ref{fig:skill-policy-examples}).
Across environments, the learned rules govern task selection, evaluation
allocation, and search control. They use failures to choose new probes, move
from cheap tests to confirmation only when warranted, and preserve or combine
changes according to accumulated evidence. Together, the intervention,
execution-trace, and skill analyses answer \textbf{Q4}: autonomy lets the
evolver request the campaign behavior it needs, while meta-evolution turns
that freedom into a reusable way of organizing improvement.

\section{Conclusion}
\label{sec:conclusion}
Agent evolvers have automated the design of harnesses, but not the broader process
that designs them. \method{} automates that process too: the evolver runs its
own improvement campaign in a free space, and meta-evolution learns how to run it from the
outcomes. Across four environments, the automated
process uses a self-controlled loop, yet matches or exceeds hand-designed
evolvers' performance. Beyond autonomy, the learned process improves
with experience. What it carries is a way of running
an improvement campaign rather than domain knowledge. Treating the
optimization process as a learnable agent capability, rather than a fixed piece
of engineering, lets agent improvement advance with experience and with the
models themselves.

\clearpage
\subsection*{AI Use Statement}
Large language models are the object of study in this work. Every evolver,
worker, simulated user, and judge is a Claude model, and the evolver agent is
Claude Code. Separately, the authors used LLM assistants to edit prose, check
consistency between tables and text, and write figure code. The authors
designed the experiments, checked every reported number against the run
artifacts, and take full responsibility for the content.

\subsection*{Ethics Statement}
Meta-evolving an agent harness can amplify errors, consume substantial compute,
and make the source of a behavioral change difficult to identify. \method{}
limits nesting, retains all evaluation records, fixes model capability, and
separates candidate generation from acceptance. These controls improve
auditability, but they do not make autonomous improvement safe for high-stakes
deployment, where domain-specific oversight remains necessary.

\subsection*{Reproducibility Statement}
Each run stores the installed environment, evaluated target, evolver
instructions, execution trace, and raw evaluation results. The appendix reports
the benchmark definitions, task splits, base agents, and model configurations.
Additional implementation and reproducibility details are available in the
anonymous code submission.

\bibliography{iclr2027_conference}

@article{terminalbench2026,
  title        = {Terminal-Bench: Benchmarking Agents on Hard, Realistic Tasks in Command Line Interfaces},
  author       = {Merrill, Mike A. and Shaw, Alexander G. and Carlini, Nicholas and Li, Boxuan and Raj, Harsh and others},
  journal      = {arXiv preprint arXiv:2601.11868},
  year         = {2026}
}

@article{bilevelAutoresearch2026,
  title        = {Bilevel Autoresearch: Meta-Autoresearching Itself},
  author       = {Qu, Yaonan and Lu, Meng},
  journal      = {arXiv preprint arXiv:2603.23420},
  year         = {2026}
}

@misc{arcagi22025,
      title={ARC-AGI-2: A New Challenge for Frontier AI Reasoning Systems}, 
      author={Francois Chollet and Mike Knoop and Gregory Kamradt and Bryan Landers and Henry Pinkard},
      year={2026},
      eprint={2505.11831},
      archivePrefix={arXiv},
      primaryClass={cs.AI},
      url={https://arxiv.org/abs/2505.11831}, 
}

@article{continualHarness2026,
  title        = {Continual Harness: Online Adaptation for Self-Improving Foundation Agents},
  author       = {Karten, Seth and Zhang, Joel and Upaa, Jr, Tersoo and Feng, Ruirong and Li, Wenzhe and Shi, Chengshuai and Jin, Chi and Vodrahalli, Kiran},
  journal      = {arXiv preprint arXiv:2605.09998},
  year         = {2026}
}

@article{lifeHarness2026,
  title        = {Adapting the Interface, Not the Model: Runtime Harness Adaptation for Deterministic {LLM} Agents},
  author       = {Xu, Tianshi and Wen, Huifeng and Li, Meng},
  journal      = {arXiv preprint arXiv:2605.22166},
  year         = {2026}
}

@article{livingHarness2026,
  title        = {Living-Harness Is an Interactive-Agent Evolver},
  author       = {Du, Yuetian and Wang, Yucheng and Xu, He and Xu, Jiexu and Tan, Shanwen and Zhao, Bing and Yang, Boyu and Xu, Zhijie and Kong, Ming and Wei, Hu and Liu, Jie and Zhu, Qiang},
  journal      = {arXiv preprint arXiv:2607.26598},
  year         = {2026}
}

@article{skilloptLite2026,
  title        = {SkillOpt-Lite: Better and Faster Agent Self-evolution via One Line of Vibe},
  author       = {Shen, Yifei and Li, Bo and Zhang, Xinjie},
  journal      = {arXiv preprint arXiv:2607.03451},
  year         = {2026}
}

@article{tthe2026,
  title        = {{TTHE}: Test-Time Harness Evolution},
  author       = {Nie, Jun and Zhang, Yonggang and Song, Jun and Cai, Qianshu and Yu, Dahai and Guo, Yike and Tian, Xinmei and Han, Bo},
  journal      = {arXiv preprint arXiv:2607.08124},
  year         = {2026}
}

@article{harnessCompass2026,
  title        = {HarnessCompass: Guiding Automatic Harness Evolution toward Generalizable and Effective Agent Harnesses},
  author       = {Zhang, Luan and Zhou, Ruochen and Song, Dandan and Chen, Zhengyu and Tian, Yuhang and Yang, Jun and Ma, Huipeng and Li, Chenhao and Feng, Guangyuan and Li, Xudong and Jin, Yizhou and Xu, Yan},
  journal      = {arXiv preprint arXiv:2608.01918},
  year         = {2026}
}

@article{agrawal2025gepa,
  title        = {{GEPA}: Reflective Prompt Evolution Can Outperform Reinforcement Learning},
  author       = {Agrawal, Lakshya A. and Tan, Shangyin and Soylu, Dilara and Ziems, Noah and Khare, Rishi and Opsahl-Ong, Krista and Singhvi, Arnav and Shandilya, Herumb and Ryan, Michael J. and Jiang, Meng and Potts, Christopher and Sen, Koushik and Dimakis, Alexandros G. and Stoica, Ion and Klein, Dan and Zaharia, Matei and Khattab, Omar},
  journal      = {arXiv preprint arXiv:2507.19457},
  year         = {2025}
}

@software{sharma2025openevolve,
  title = {OpenEvolve: an open-source evolutionary coding agent},
  author = {Asankhaya Sharma},
  year = {2025},
  publisher = {GitHub},
  url = {https://github.com/algorithmicsuperintelligence/openevolve}
}

@article{liu2026evox,
  title        = {{EvoX}: Meta-Evolution for Automated Discovery},
  author       = {Liu, Shu and Agarwal, Shubham and Maheswaran, Monishwaran and Cemri, Mert and Li, Zhifei and Mang, Qiuyang and Naren, Ashwin and Boneh, Ethan and Cheng, Audrey and Pan, Melissa Z. and Du, Alexander and Keutzer, Kurt and Cheung, Alvin and Dimakis, Alexandros G. and Sen, Koushik and Zaharia, Matei and Stoica, Ion},
  journal      = {arXiv preprint arXiv:2602.23413},
  year         = {2026}
}

@article{zhang2026hyperagents,
  title        = {HyperAgents},
  author       = {Zhang, Jenny and Zhao, Bingchen and Yang, Wannan and Foerster, Jakob and Clune, Jeff and Jiang, Minqi and Devlin, Sam and Shavrina, Tatiana},
  journal      = {arXiv preprint arXiv:2603.19461},
  year         = {2026}
}

@article{shi2026tau,
  title        = {$\tau$-Knowledge: Evaluating Conversational Agents over Unstructured Knowledge},
  author       = {Shi, Quan and Zytek, Alexandra and Razavi, Pedram and Narasimhan, Karthik and Barres, Victor},
  journal      = {arXiv preprint arXiv:2603.04370},
  year         = {2026}
}

@article{autoHarness2026,
  title        = {{AutoHarness}: Improving {LLM} Agents by Automatically Synthesizing a Code Harness},
  author       = {Lou, Xinghua and L{\'a}zaro-Gredilla, Miguel and Dedieu, Antoine and Wendelken, Carter and Lehrach, Wolfgang and Murphy, Kevin P.},
  journal      = {arXiv preprint arXiv:2603.03329},
  year         = {2026}
}

@article{metaHarness2026,
  title        = {{Meta-Harness}: End-to-End Optimization of Model Harnesses},
  author       = {Lee, Yoonho and Nair, Roshen and Zhang, Qizheng and Lee, Kangwook and Khattab, Omar and Finn, Chelsea},
  journal      = {arXiv preprint arXiv:2603.28052},
  year         = {2026}
}

@article{agenticHarnessEngineering2026,
  title        = {Agentic Harness Engineering: Observability-Driven Automatic Evolution of Coding-Agent Harnesses},
  author       = {Lin, Jiahang and Liu, Shichun and Pan, Chengjun and Lin, Lizhi and Dou, Shihan and others},
  journal      = {arXiv preprint arXiv:2604.25850},
  year         = {2026}
}

@article{metaContextEngineering2026,
  title        = {Meta Context Engineering via Agentic Skill Evolution},
  author       = {Ye, Haoran and He, Xuning and Arak, Vincent and Dong, Haonan and Song, Guojie},
  journal      = {arXiv preprint arXiv:2601.21557},
  year         = {2026}
}

@misc{aide22026,
      title={Recursive self-improvement of AI research agents}, 
      author={Dhruv Srikanth and Bingchen Zhao and Dixing Xu and Yuxiang Wu and Zhengyao Jiang},
      year={2026},
      eprint={2609.26457},
      archivePrefix={arXiv},
      primaryClass={cs.AI},
      url={https://arxiv.org/abs/2609.26457}, 
}

@inproceedings{promptBreeder2023,
  title        = {Promptbreeder: Self-Referential Self-Improvement via Prompt Evolution},
  author       = {Fernando, Chrisantha and Banarse, Dylan Sunil and Michalewski, Henryk and Osindero, Simon and Rockt{\"a}schel, Tim},
  booktitle    = {Proceedings of the 41st International Conference on Machine Learning (ICML)},
  series       = {Proceedings of Machine Learning Research},
  volume       = {235},
  pages        = {13481--13544},
  year         = {2024}
}

@inproceedings{stop2024,
  title        = {Self-Taught Optimizer ({STOP}): Recursively Self-Improving Code Generation},
  author       = {Zelikman, Eric and Lorch, Eliana and Mackey, Lester and Kalai, Adam Tauman},
  booktitle    = {First Conference on Language Modeling (COLM)},
  year         = {2024}
}

@inproceedings{godelAgent2025,
  title        = {G{\"o}del Agent: A Self-Referential Agent Framework for Recursively Self-Improvement},
  author       = {Yin, Xunjian and Wang, Xinyi and Pan, Liangming and Lin, Li and Wan, Xiaojun and Wang, William Yang},
  booktitle    = {Proceedings of the 63rd Annual Meeting of the Association for Computational Linguistics (Volume 1: Long Papers)},
  pages        = {27890--27913},
  year         = {2025}
}

@inproceedings{darwinGodelMachine2025,
  title        = {Darwin G{\"o}del Machine: Open-Ended Evolution of Self-Improving Agents},
  author       = {Zhang, Jenny and Hu, Shengran and Lu, Cong and Lange, Robert and Clune, Jeff},
  booktitle    = {The Fourteenth International Conference on Learning Representations (ICLR)},
  year         = {2026}
}

@inproceedings{hu2025adas,
  title        = {Automated Design of Agentic Systems},
  author       = {Hu, Shengran and Lu, Cong and Clune, Jeff},
  booktitle    = {The Thirteenth International Conference on Learning Representations (ICLR)},
  year         = {2025}
}

@inproceedings{zhang2025aflow,
  title        = {{AFlow}: Automating Agentic Workflow Generation},
  author       = {Zhang, Jiayi and Xiang, Jinyu and Yu, Zhaoyang and Teng, Fengwei and Chen, Xionghui and others},
  booktitle    = {The Thirteenth International Conference on Learning Representations (ICLR)},
  year         = {2025}
}

@article{novikov2025alphaevolve,
  title        = {{AlphaEvolve}: A Coding Agent for Scientific and Algorithmic Discovery},
  author       = {Novikov, Alexander and V{\~u}, Ng{\^a}n and Eisenberger, Marvin and Dupont, Emilien and Huang, Po-Sen and others},
  journal      = {arXiv preprint arXiv:2506.13131},
  year         = {2025}
}

@article{robeyns2025sica,
  title        = {A Self-Improving Coding Agent},
  author       = {Robeyns, Maxime and Szummer, Martin and Aitchison, Laurence},
  journal      = {arXiv preprint arXiv:2504.15228},
  year         = {2025}
}

@article{jiang2025aide,
  title        = {{AIDE}: {AI}-Driven Exploration in the Space of Code},
  author       = {Jiang, Zhengyao and Schmidt, Dominik and Srikanth, Dhruv and Xu, Dixing and Kaplan, Ian and others},
  journal      = {arXiv preprint arXiv:2502.13138},
  year         = {2025}
}

@article{arcagi32026,
  title        = {{ARC-AGI-3}: A New Challenge for Frontier Agentic Intelligence},
  author       = {{ARC Prize Foundation}},
  journal      = {arXiv preprint arXiv:2603.24621},
  year         = {2026}
}

@inproceedings{liu2026skydiscover,
  author    = {Liu, Shu and Cemri, Mert and Agarwal, Shubham and Krentsel, Alexander and Naren, Ashwin and Mang, Qiuyang and Li, Zhifei and Gupta, Akshat and Maheswaran, Monishwaran and Cheng, Audrey and Pan, Melissa and Boneh, Ethan and Ramchandran, Kannan and Sen, Koushik and Zaharia, Matei and Dimakis, Alexandros G. and Stoica, Ion},
  title     = {SkyDiscover: A Flexible, Adaptive Framework for AI-Driven Scientific and Algorithmic Discovery},
  booktitle = {Proceedings of the ACM Conference on AI and Agentic Systems},
  series    = {CAIS '26},
  year      = {2026},
  pages     = {1223--1227},
  publisher = {Association for Computing Machinery},
  doi       = {10.1145/3786335.3813221},
  url       = {https://doi.org/10.1145/3786335.3813221}
}
\bibliographystyle{iclr2027_conference}

\newpage
\appendix
\section*{Appendix}
\section{Implementation Details}
\label{app:implementation}
\paragraph{Anonymous code.}

We instantiate the evolver with Claude Code and a single instruction file,
\texttt{SKILL.md}. One call to \texttt{evolve.sh} installs an environment in a
fresh run directory and invokes this agent once. There is no framework-level
evolution loop around that invocation. The seed skill contains only general
instructions for working from evidence and leaving the strongest target
behind; it does not prescribe an iteration count, branching structure,
curriculum, evaluation schedule, or performance-based stopping rule.

Each installed environment realizes Equation~\ref{eq:environment} through four
visible components. \texttt{ENVIRONMENT.md} records the goal, data, constraints,
and interpretation of the returned artifacts. The file \texttt{.target}
identifies the editable surface \(x_0\), while \texttt{run.sh} implements
\(R_{\mathcal{E}}\). An initially empty \texttt{workspace-<env>/} stores the
candidates, artifacts, and notes produced during evolution. The environment
files remain fixed throughout a campaign; only the declared target is editable.

The call \(R_{\mathcal{E}}(x,u)\) is implemented in place. The evolver installs
candidate \(x\) at the path declared by \texttt{.target} and supplies \(u\)
through the arguments of \texttt{run.sh}. Each call records its measurements
and traces in an artifact directory. Evaluation always reads the declared edit
surface, ensuring that the candidate under evaluation is the candidate being
evolved.

The run directory is the complete interface between environment and evolver.
It is self-contained, records the skill and model used for the run, and
preserves the evaluation artifacts produced during the campaign. The evolver
may invoke \texttt{run.sh} in any order or configuration, launch independent
branches, and decide when enough evidence has been collected. The framework
enforces the environment boundary, but contributes no search schedule.

The meta-environment uses exactly the same layout. Its declared target is a
candidate \texttt{SKILL.md}, and its \texttt{run.sh} launches that skill as the
evolver in a fresh worker environment. A meta-level execution returns the
complete inner run directory, including the worker produced, its evaluations,
and the inner trace, rather than reducing the campaign to a fixed scalar. The
outer evolver chooses which worker environment to launch and how many inner
runs to collect. Meta-environments are excluded from the inner environment
catalogue, bounding execution to the two levels described above.

\paragraph{Hardware and campaign budgets.}
Standalone worker-evolution runs receive a 12-hour wall-clock budget on one
128-CPU machine. The environment
permits at most 24 concurrent task attempts, regardless of how the evolver
divides them among evaluation calls.

Meta-evolution receives a 24-hour outer-campaign budget on a Slurm cluster
with five worker nodes. The meta-environment
submits each requested inner campaign as an independent one-node Slurm job, and
\(\mathcal{C}_{\mathrm{m}}\) permits at most five inner campaigns to run
simultaneously. We impose no separate wall-clock limit on an inner evolver
within meta-evolution: deciding how many inner campaigns to launch, how long
to let them run, and how to coordinate asynchronous completions is part of the
meta-evolver's campaign control.

\section{Dataset Details}
\begin{table*}[t]
\caption{Benchmark datasets, metrics, evaluation protocols, and base agents.}
\label{tab:dataset}
\begin{center}
\small
\setlength{\tabcolsep}{4pt}
\begin{tabular}{@{}p{0.13\textwidth}p{0.34\textwidth}p{0.22\textwidth}p{0.23\textwidth}@{}}
\toprule
Benchmark & Dataset and primary metric & Evolution / held-out protocol
& Base agent \\ \midrule
\(\tau^3\)-bench &
97 interactive \texttt{banking\_knowledge} conversations with tools and a
domain policy; mean task pass rate and pass@3. &
68 evolution tasks and 29 disjoint held-out tasks. &
Vendored \(\tau^3\) customer-service agent with Claude Opus 4.8. \\[3pt]
ARC-AGI-2 &
Few-shot visual grid-transformation tasks that test abstract rule induction;
mean task accuracy and pass@3. &
56 evolution tasks and 111 disjoint validation tasks. &
Base ARC solver backed by Claude 4.7 Opus. \\[3pt]
ARC-AGI-3 &
Interactive abstract-reasoning games; Relative Human Action Efficiency
(RHAE) score. &
17 evolution games and 8 disjoint held-out games. &
Base ARC-AGI-3 agent backed by Claude Opus 5. \\[3pt]
Terminal-Bench 2.1 &
Containerized command-line tasks covering software, systems, security, and
scientific workflows; mean task pass rate and pass@3. &
Evolution tasks are hidden from final evaluation; 61 held-out tasks. &
Unmodified deep agent with Claude Opus 4.8. \\
\bottomrule
\end{tabular}
\end{center}
\end{table*}

\paragraph{\(\tau^3\)-bench.}
We use the \texttt{banking\_knowledge} domain. Each task is a tool-mediated
conversation with a simulated customer. The agent must retrieve the relevant
bank policy, take any required actions, and communicate the correct resolution.
The grader assigns a binary reward after the conversation. The full domain
contains 97 tasks; our fixed split places 68 tasks in the evolution partition
and 29 in the held-out partition.

\paragraph{ARC-AGI-2.}
ARC-AGI-2 consists of few-shot visual reasoning tasks represented as
input--output grid transformations \citep{arcagi22025}. A solver must infer the
latent transformation from the demonstrations and apply it to a new input. We
use 56 tasks for evolution and a disjoint set of 111 held-out tasks.
Meta-evolution may use only the 56-task evolution partition. We report task
accuracy on the held-out partition.

\paragraph{ARC-AGI-3.}
ARC-AGI-3 evaluates agents in interactive abstract-reasoning games
\citep{arcagi32026}. We use 17 games for evolution and reserve 8 disjoint games
for held-out evaluation. The agent interacts with each game through the
declared environment interface, while the game definitions and evaluator
remain fixed. We report Relative Human Action Efficiency (RHAE). Each
completed level receives the squared ratio of human-baseline actions to agent
actions, subject to the benchmark's cap for faster-than-human play;
uncompleted levels receive zero. Level scores are linearly weighted so later
levels count more, each game is capped by the weighted fraction of levels
completed, and RHAE averages the resulting game scores. We report it as a
percentage, with higher values indicating more complete and action-efficient
play.

\paragraph{Terminal-Bench 2.1.}
Terminal-Bench 2.1 contains realistic command-line tasks executed in isolated
containers \citep{terminalbench2026}. Tasks include software debugging, systems
configuration, security, data processing, and scientific workflows. Each task
provides a problem statement and a test suite that verifies the final container
state. We report results on 61 held-out tasks. We report mean and pass@3.

All task partitions are created before meta-evolution. Neither the worker-level
nor the meta-level evolver may access the held-out partition. The same split is
used for all compared evolver generations within a benchmark. To ensure evolver do not break out of the working directory to find held-out set, we perform keyword search of held-out data IDs over the run directory. The evolvers consistently stay within the established run directory without attempting to hack held-out data.

\section{Configuration Details}
\paragraph{Model matching.}
Unless otherwise stated, the worker, worker-level evolver, and outer
meta-evolver use the same Claude model within an environment. The
model-capability experiment is the explicit exception. This design evaluates
self-improvement without introducing a stronger teacher model.

Each environment is installed into a fresh run directory containing its
\texttt{ENVIRONMENT.md}, evaluator, declared edit surface, and an empty
workspace. The evolver runs from that directory. It may modify the declared
target and write to its workspace; it cannot edit the environment or redirect the
evaluator to another target.

\paragraph{\(\tau^3\)-bench configuration.}
The base worker is the vendored \(\tau^3\) customer-service agent. The agent,
simulated user, and judge each use
\texttt{bedrock/global.anthropic.claude-opus-4-8}. The agent's reasoning effort
is fixed to \texttt{medium}. The editable surface includes the agent prompt,
tool-use loop, response parsing, retry behavior, and termination logic; changing
the model, reasoning capability, task data, tools, policy, or grader is out of
scope.

\paragraph{ARC-AGI-2 configuration.}
The base worker is an ARC-AGI-2 solver backed by Claude 4.7 Opus. The model and
inference configuration are held fixed across the base, seed-evolver, and
meta-evolved conditions. The evolver may modify the solver harness and its
search procedure, but not the task grids, evolution/held-out split, or
correctness evaluator. Each reported worker is produced from the same base
solver rather than inherited from the preceding condition.

\paragraph{ARC-AGI-3 configuration.}
The base worker is an ARC-AGI-3 agent backed by Claude Opus 5. The model,
reasoning configuration, game definitions, and evaluator are fixed across
conditions. The evolver may modify the declared agent harness and search
procedure but may not modify the games or their scoring implementation.

\paragraph{Terminal-Bench 2.1 configuration.}
The base worker is the unmodified deep agent. It uses
\texttt{bedrock/global.anthropic.claude-opus-4-8} with temperature
\texttt{1.0}. The editable surface includes the agent control loop, prompts,
response parsers, and terminal-session handling. Container definitions, task
statements, test suites, and verifier code remain fixed.

\section{Additional Results}
\label{app:additional-results}

\begin{figure}[!t]
  \centering
\definecolor{spSeedFill}{HTML}{F1F0ED}%
\definecolor{spSeedEdge}{HTML}{B7B5AF}%
\definecolor{spBlue}{HTML}{19517A}%
\definecolor{spBlueFill}{HTML}{EAF3F9}%
\definecolor{spBlueEdge}{HTML}{86B9DB}%
\definecolor{spInk}{HTML}{202225}%
\definecolor{spMuted}{HTML}{65686C}%
\begin{tikzpicture}[x=1cm,y=1cm]
\tikzset{
  sp seed/.style={
    draw=spSeedEdge, fill=spSeedFill, rounded corners=2pt,
    line width=0.6pt, text width=3.05cm, minimum height=1.50cm,
    inner sep=4.5pt, align=left, font=\scriptsize, text=spInk,
  },
  sp learned/.style={
    draw=spBlueEdge, fill=spBlueFill, rounded corners=2pt,
    line width=0.8pt, text width=5.25cm, minimum height=1.50cm,
    inner sep=4.5pt, align=left, font=\scriptsize, text=spInk,
  },
  sp effect/.style={
    draw=spBlue, fill=spBlue, rounded corners=2pt,
    line width=0.8pt, text width=2.65cm, minimum height=1.50cm,
    inner sep=4.5pt, align=left, font=\scriptsize, text=white,
  },
  sp head/.style={font=\scriptsize\bfseries, text=spMuted, align=center},
  sp arrow/.style={->, >=stealth, line width=0.8pt, draw=spBlue!75},
}

\node[sp head] at (1.75,0.15) {Common seed disposition};
\node[sp head, text=spBlue] at (6.72,0.15) {Policy learned by meta-evolution};
\node[sp head, text=spBlue] at (11.40,0.15) {Realized advantage};

\node[sp seed] (seed1) at (1.75,-0.95) {
  \textbf{Task selection}\\[1pt]
  \textit{``Some failure you have not explained.''}
};
\node[sp learned] (learn1) at (6.72,-0.95) {
  \textbf{\(\tau^3\)-bench skill}\\[1pt]
  \textit{``Read transcripts, tool calls, and grader output; group failures
  by cause; work on the largest group.''}
};
\node[sp effect] (effect1) at (11.40,-0.95) {
  \textbf{Adaptive curriculum}\\[2pt]
  Target the largest failure cluster.
};
\draw[sp arrow] (seed1.east) -- (learn1.west);
\draw[sp arrow] (learn1.east) -- (effect1.west);

\node[sp seed] (seed2) at (1.75,-2.68) {
  \textbf{Candidate promotion}\\[1pt]
  \textit{``Your next attempt starts from your best.''}
};
\node[sp learned] (learn2) at (6.72,-2.68) {
  \textbf{ARC-AGI-3 skill}\\[1pt]
  Use per-item signs, not only the mean; re-evaluate the nominee; keep the
  accumulated version over the luckiest draw.
};
\node[sp effect] (effect2) at (11.40,-2.68) {
  \textbf{Regression control}\\[2pt]
  Keep supported gains, not a noisy maximum.
};
\draw[sp arrow] (seed2.east) -- (learn2.west);
\draw[sp arrow] (learn2.east) -- (effect2.west);
\end{tikzpicture}
  \caption{Campaign rules added to the seed skill by meta-evolution.}
  \label{fig:skill-policy-examples}
\end{figure}
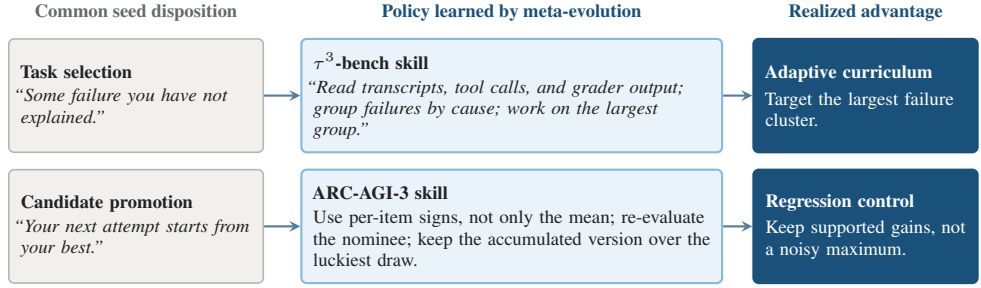

\paragraph{Complete numerical results for the main figures.}
Table~\ref{tab:frozen-evolver} reports the full fresh-worker comparison, and
Table~\ref{tab:evolver-model-scaling} reports the full model-capability study.
\begin{table*}[t]
\caption{Fresh-worker evaluation of seed, no-data, and meta-evolved evolvers.}
\label{tab:frozen-evolver}
\begin{center}
\footnotesize
\setlength{\tabcolsep}{2pt}
\begin{tabular}{@{}llccccc@{}}
\toprule
Dataset & Metric & Base worker & Seed skill & No-data LLM
& Meta-evolved & \(\Delta\) \\
\midrule
\multirow{2}{*}{\(\tau^3\)-bench}
& Mean & 33.3 {\scriptsize\(\pm 5.3\)}
& 33.3 {\scriptsize\(\pm 5.3\)}
& 37.9 {\scriptsize\(\pm 6.0\)}
& \textbf{47.1 {\scriptsize\(\pm 8.7\)}} & \(+13.8\) \\
& Pass@3 & 48.3 & 48.3 & 58.6 & \textbf{62.1} & \(+13.8\) \\
\midrule
\multirow{2}{*}{ARC-AGI-2}
& Mean & 55.4 {\scriptsize\(\pm 4.0\)}
& 56.5 {\scriptsize\(\pm 1.4\)}
& 56.8 {\scriptsize\(\pm 3.1\)}
& \textbf{59.8 {\scriptsize\(\pm 1.9\)}} & \(+3.3\) \\
& Pass@3 & 75.0 & 74.8 & 74.8 & \textbf{81.1} & \(+6.3\) \\
\midrule
ARC-AGI-3
& RHAE & 61.0 {\scriptsize\(\pm 1.1\)}
& 61.7 {\scriptsize\(\pm 4.3\)}
& 64.5 {\scriptsize\(\pm 5.1\)}
& \textbf{70.3 {\scriptsize\(\pm 3.2\)}} & \(+8.6\) \\
\midrule
\multirow{2}{*}{Terminal-Bench 2.1}
& Mean & 80.9 {\scriptsize\(\pm 5.0\)}
& 81.4 {\scriptsize\(\pm 2.5\)}
& 78.7 {\scriptsize\(\pm 3.3\)}
& \textbf{83.1 {\scriptsize\(\pm 4.1\)}} & \(+1.7\) \\
& Pass@3 & 88.5 & 86.9 & 86.9 & \textbf{90.2} & \(+3.3\) \\
\bottomrule
\end{tabular}
\end{center}
\end{table*}

\begin{table*}[t]
\caption{Effect of evolver-model capability on meta-evolution for
\(\tau^3\)-bench.}
\label{tab:evolver-model-scaling}
\begin{center}
\scriptsize
\setlength{\tabcolsep}{9pt}
\begin{tabular}{@{}lccc@{}}
\toprule
Meta-evolution model & Evolution & Held-out mean & Pass@3 \\
\midrule
No meta-evolution & 35.0 & 33.3 {\scriptsize\(\pm 5.3\)} & 48.3 \\
\midrule
Haiku 4.5  & 52.2 & 32.2 {\scriptsize\(\pm 8.7\)} & 48.3 \\
Sonnet 4.6 & 56.4 & 47.1 {\scriptsize\(\pm 4.0\)} & 62.1 \\
Opus 4.6   & 48.0 & 40.2 {\scriptsize\(\pm 2.0\)} & 55.2 \\
Opus 4.7   & \textbf{59.5} & 49.4 {\scriptsize\(\pm 8.0\)}
& \textbf{69.0} \\
Opus 4.8   & 51.5 & \textbf{51.7 {\scriptsize\(\pm 6.0\)}}
& \textbf{69.0} \\
\bottomrule
\end{tabular}
\end{center}
\vspace{-5mm}
\end{table*}

\paragraph{Fixed campaign-decision ablation.}
Table~\ref{tab:fixed-u} gives the complete numerical results behind
Table~\ref{tab:fixed-u-main}, including evaluation variation and Pass@3.
\begin{table}[!t]
\caption{Performance and total token use when individual campaign decisions
are fixed on \(\tau^3\)-bench.}
\label{tab:fixed-u}
\begin{center}
\scriptsize
\setlength{\tabcolsep}{4pt}
\begin{tabular}{@{}lcccc@{}}
\toprule
& \textbf{\shortstack{Evolver-controlled\\\(u\)}}
& \shortstack{Always\\full set}
& \shortstack{Concurrency\\\(=24\)}
& \shortstack{No task\\timeout} \\
\midrule
Held-out mean
& \textbf{47.1 \(\pm 8.7\)}
& 39.1 \(\pm 4.3\)
& 43.7 \(\pm 4.3\)
& 44.8 \(\pm 3.5\) \\
Pass@3
& \textbf{62.1}
& 58.6
& 58.6
& 55.2 \\
Total tokens
& \textbf{1.19B}
& 0.89B
& 1.78B
& 1.53B \\
\bottomrule
\end{tabular}
\end{center}
\vspace{-5mm}
\end{table}

\paragraph{Campaign wall-clock time.}
\begin{table}[!t]
\caption{Wall-clock hours by method and benchmark. The \method{} row reports
its 24-hour outer-campaign budget.}
\label{tab:campaign-wall-clock}
\begin{center}
\small
\setlength{\tabcolsep}{8pt}
\begin{tabular}{@{}lrrrr@{}}
\toprule
Method
& \(\tau^3\)-bench
& ARC-AGI-2
& ARC-AGI-3
& TB 2.1 \\
\midrule
GEPA & 11.0 & 11.9 & 30.1 & 32.1 \\
OpenEvolve & 9.3 & 16.5 & 36.0 & 25.7 \\
EvoX & 6.9 & 13.0 & 33.5 & 25.8 \\
HyperAgents & 11.9 & 13.4 & 23.8 & 29.7 \\
\method{} meta-evolution & 24.0 & 24.0 & 24.0 & 24.0 \\
\bottomrule
\end{tabular}
\end{center}
\vspace{-5mm}
\end{table}
Wall-clock times for all compared campaigns are
reported in Appendix Table~\ref{tab:campaign-wall-clock}.
\begin{table*}[t]
\caption{Complete cross-model transfer results for frozen worker harnesses.}
\label{tab:tau-cross-model}
\begin{center}
\scriptsize
\setlength{\tabcolsep}{3.2pt}
\begin{tabular}{@{}llcccccc@{}}
\toprule
& & \multicolumn{2}{c}{Native source model}
& \multicolumn{2}{c}{Sonnet 4.6}
& \multicolumn{2}{c}{Haiku 4.5} \\
\cmidrule(lr){3-4}\cmidrule(lr){5-6}\cmidrule(lr){7-8}
Dataset & Evolution method
& Mean & Pass@3 & Mean & Pass@3 & Mean & Pass@3 \\
\midrule
\multirow{6}{*}{\(\tau^3\)-bench}
& Base agent
& 33.3 {\scriptsize\(\pm 5.3\)} & 48.3
& 40.2 {\scriptsize\(\pm 8.0\)} & 55.2
& 17.2 {\scriptsize\(\pm 6.0\)} & 34.5 \\
& GEPA
& 35.6 {\scriptsize\(\pm 4.0\)} & 51.7
& 36.8 {\scriptsize\(\pm 8.0\)} & 51.7
& 9.2 {\scriptsize\(\pm 4.0\)} & 20.7 \\
& OpenEvolve
& 35.6 {\scriptsize\(\pm 4.0\)} & 55.2
& 35.6 {\scriptsize\(\pm 5.3\)} & 62.1
& 14.9 {\scriptsize\(\pm 5.3\)} & 31.0 \\
& EvoX
& 36.8 {\scriptsize\(\pm 2.0\)} & 58.6
& 39.1 {\scriptsize\(\pm 8.0\)} & 44.8
& 11.5 {\scriptsize\(\pm 5.3\)} & 24.1 \\
& HyperAgents
& 36.8 {\scriptsize\(\pm 5.3\)} & 58.6
& 35.6 $\pm{4.0}$ & 48.3 & 14.9 $\pm{3.0}$ & 31.0 \\
& \textbf{\method{}}
& \textbf{51.7 {\scriptsize\(\pm 6.0\)}} & \textbf{69.0}
& \textbf{47.1 {\scriptsize\(\pm 7.2\)}} & \textbf{65.5}
& \textbf{20.7 {\scriptsize\(\pm 3.5\)}} & \textbf{37.9} \\
\midrule
\multirow{6}{*}{ARC-AGI-2}
& Base agent
& 55.4 {\scriptsize\(\pm 4.0\)} & 75.0
& 39.7 $\pm{4.4}$ & 50.0 & 4.8 $\pm{1.5}$ & 10.0 \\
& GEPA
& 58.0 {\scriptsize\(\pm 5.7\)} & 76.4
& 38.4 $\pm{4.2}$ & 53.5 & 5.7 $\pm{1.8}$ & 11.7 \\
& OpenEvolve
& 58.6 {\scriptsize\(\pm 5.0\)} & 79.1
& 44.1 $\pm{3.8}$ & 57.3 & 4.8 $\pm{1.9}$& 7.2 \\
& EvoX
& 64.0 {\scriptsize\(\pm 6.4\)} & 80.0
& 43.8 $\pm{4.2}$ & 55.9 & 5.4 $\pm{1.9}$& 10.0 \\
& HyperAgents
& 59.6 {\scriptsize\(\pm 4.2\)} & 77.8
& 40.1 $\pm{4.2}$ & 55.2 & 6.3 $\pm{1.9}$& 11.7 \\
& \textbf{\method{}}
& \textbf{69.1 {\scriptsize\(\pm 5.4\)}} & \textbf{86.5}
& \textbf{46.3} {\scriptsize\(\pm 1.9\)} & \textbf{62.2}
& \textbf{9.0} {\scriptsize\(\pm 1.6\)} & \textbf{15.3} \\
\bottomrule
\end{tabular}
\end{center}
\end{table*}

\begin{table}[t]
\caption{Experimental questions Q1--Q5 and the main-text evidence that
answers each.}
\label{tab:questions}
\begin{center}
\small
\begin{tabular}{
  >{\raggedright\arraybackslash}p{0.33\linewidth}
  >{\raggedright\arraybackslash}p{0.61\linewidth}}
\toprule
Question & Evidence \\ \midrule
\textbf{Q1: Can a learned autonomous campaign achieve strong performance
without a prescribed loop?} &
\method{} improves the primary held-out metric by \(13.6\) points on average
over the base worker, outperforms the expert-designed evolvers on three of
four benchmarks, and matches or outperforms the meta-evolution baselines EvoX
and HyperAgents. On Terminal-Bench 2.1 it has the best pass@3 but trails GEPA
in mean score (Table~\ref{tab:main}). \\ \addlinespace[3pt]
\textbf{Q2: Does meta-evolution turn campaign autonomy into a stronger
evolver?} &
On fresh workers, the seed skill gains only \(0.6\) points on average across
the four mean metrics. The frozen meta-evolved skill adds a further \(6.9\)
points, improves every available pass@3 metric, and outperforms the no-data
LLM rewrite in every environment (Figure~\ref{fig:frozen-evolver}). \\ \addlinespace[3pt]
\textbf{Q3: Does the resulting evolution skill generalize beyond its source
environment and model?} &
Frozen skills transfer positively on most target metrics, but both
single-source skills reduce the Terminal-Bench 2.1 mean; joint meta-evolution
on \(\tau^3\)-bench and Terminal-Bench 2.1 improves every target metric over
the seed (Table~\ref{tab:cross-environment}). Evolved worker harnesses also
transfer across models: run unchanged with Sonnet 4.6 and Haiku 4.5,
\method{} keeps a \(5.3\)-point average gain over the base harness
(Figure~\ref{fig:cross-model-transfer}a). \\ \addlinespace[3pt]
\textbf{Q4: How does the evolver use campaign autonomy as evolution unfolds?}
&
Fixing any single evaluation decision (always the full set, concurrency
\(=24\), or no timeout) yields a weaker worker, and the last two also use more
tokens (Table~\ref{tab:fixed-u-main}). The ARC-AGI-2 trace shows the evolver
changing evaluation resolution, opening branches, combining discoveries, and
selecting the final worker as evidence accumulates
(Figure~\ref{fig:arc2-campaign-behaviors}); meta-evolved skills turn the
seed's dispositions into operational rules
(Figure~\ref{fig:skill-policy-examples}). \\ \addlinespace[3pt]
\textbf{Q5: What model capability is required to exercise campaign autonomy
effectively?} &
With Haiku 4.5 as the evolver, meta-evolution improves neither held-out
metric. From Sonnet 4.6 onward it consistently improves held-out performance,
with the strongest Opus models producing the best workers, although Sonnet
4.6 outperforms Opus 4.6 (Table~\ref{tab:evolver-model-scaling-main}). \\
\bottomrule
\end{tabular}
\end{center}
\end{table}

\paragraph{Interpretation.}
The end-to-end results show that evolver-controlled campaigns improve the
complete system across all four environments. Fresh-worker tests isolate
improvement stored in the evolution skill, while cross-environment evaluation
tests whether that improvement transfers. The fixed-\(u\) study isolates
several environment-facing decisions, and the campaign trace shows that the
evolver also adapts search structure.

\section{Additional Related Work}
\label{app:related-work}
\paragraph{Broader harness evolution and self-improvement.}
ADAS and AFlow search over coded agent designs and workflows
\citep{hu2025adas,zhang2025aflow}. Continual Harness adapts prompts, tools,
skills, and memory during interaction \citep{continualHarness2026};
Life-Harness, Living-Harness, and SkillOpt-Lite retain reusable runtime or
procedural improvements
\citep{lifeHarness2026,livingHarness2026,skilloptLite2026}; and AutoHarness and
Agentic Harness Engineering search over code and workflow surfaces
\citep{autoHarness2026,agenticHarnessEngineering2026}. AlphaEvolve provides a
general evolutionary coding system with a prescribed evaluation loop
\citep{novikov2025alphaevolve}. Self-improving systems have also widened the
improver's edit surface: PromptBreeder evolves mutation prompts
\citep{promptBreeder2023}; STOP applies a program improver to its own source
\citep{stop2024}; and G\"odel Agent, the Darwin G\"odel Machine, and SICA let
coding agents modify their own code
\citep{godelAgent2025,darwinGodelMachine2025,robeyns2025sica}. Bilevel
Autoresearch and Meta Context Engineering improve mechanisms or skills using
downstream outcomes
\citep{bilevelAutoresearch2026,metaContextEngineering2026}. Autonomous
research agents such as AIDE choose their own experiments while optimizing a
task artifact \citep{jiang2025aide}; \method{} instead learns the reusable
skill that organizes future improvement campaigns.

\paragraph{Transferring products and improvement procedures.}
HarnessCompass, Meta-Harness, Agentic Harness Engineering, and the Darwin
G\"odel Machine evaluate whether final evolved harnesses or agents transfer to
new tasks or models
\citep{harnessCompass2026,metaHarness2026,agenticHarnessEngineering2026,
darwinGodelMachine2025}. TTHE instead adapts executable harnesses on the
evaluation tasks themselves \citep{tthe2026}. These studies primarily transfer
the \emph{product} of improvement. STOP transfers an evolved program improver
across small algorithmic tasks \citep{stop2024}. HyperAgents transfers
meta-level improvement mechanisms, and AIDE\(^2\) evaluates an evolved
optimizer beyond the tasks that produced it
\citep{zhang2026hyperagents,aide22026}. \method{} freezes the meta-evolved
evolution skill before it sees the target environment and installs it in an
evolver operating on a fresh worker. The target supplies its own data, model,
execution function, and resource constraints; no source worker, search branch,
or execution artifact is retained. This tests whether the learned improvement
process itself generalizes, rather than only whether its previously evolved
product does.

\clearpage
\section{Concrete Example for Worker and Meta-Evolver Environments}
\label{app:environment-example}
For both worker and meta-evolver environments, \texttt{ENVIRONMENT.md}
specifies the goal, editable target, available data, resource limits, and edit
constraints. The worker environment and top-level meta-environment each
provide their own file. Table~\ref{tab:environment-levels} compares the two
contracts for ARC-AGI-3.

\begin{table}[H]
\centering
\caption{\textbf{ARC-AGI-3 worker and meta-evolver environments.}
The shared contract fields hide different targets and units of evaluation.}
\label{tab:environment-levels}
\footnotesize
\setlength{\tabcolsep}{4pt}
\renewcommand{\arraystretch}{1.08}
\begin{tabular}{@{}>{\raggedright\arraybackslash}p{0.16\linewidth}
                    >{\raggedright\arraybackslash}p{0.39\linewidth}
                    >{\raggedright\arraybackslash}p{0.39\linewidth}@{}}
\toprule
Contract field
& Worker environment: \texttt{ARC-AGI-3}
& Meta-evolver environment: \texttt{meta\_chain} \\
\midrule
Goal
& Improve the game-playing agent so it completes more ARC-AGI-3 levels with
  fewer actions.
& Improve the evolution skill itself, and use it to produce strong agents in
  every installed inner environment. \\
\addlinespace
Editable target
& \texttt{agent/arc3\_agent/}: the worker's harness and game-playing loop.
& \texttt{target/SKILL.md}: the instructions that control a complete
  worker-evolution campaign. \\
\addlinespace
Held fixed
& Model, reasoning effort, token budget, game engine, games, scoring rule,
  2,000-action limit, and 7,200-second attempt limit.
& Inner environments and their evaluators. The seed's dispositions are
  preserved while its operational method is evolved. \\
\addlinespace
One evaluation
& Run a worker version on selected games and attempts:
  \texttt{./run.sh --artifacts <dir> [--tasks ...] [--attempts ...]}.
& Install the candidate skill as an inner evolver and launch a complete
  worker-evolution run:
  \texttt{./run.sh --artifacts <dir> --inner-env arc\_agi\_3}. \\
\addlinespace
Evidence
& A scalar Relative Human Action Efficiency (RHAE) mean plus levels completed,
  actions, errors, per-game results, action traces, and agent transcripts.
& No single scalar. The outer evolver reads the inner run's evaluations,
  candidate workers, final absolute quality, cost, and campaign trace. \\
\addlinespace
Evaluation controls
& Choose games, attempts, concurrency, and whole-evaluation timeout.
& Choose the inner environment and run candidate skills concurrently. \\
\addlinespace
Persistent result
& An improved worker in \texttt{agent/arc3\_agent/}, together with its
  evaluation artifacts.
& An improved evolver in \texttt{target/SKILL.md}; inner runs additionally
  retain their best workers and all nested evidence. \\
\addlinespace
Resource horizon
& Standalone comparisons allow 12 hours on one 128-CPU node, with at most 24
  concurrent task attempts.
& The outer campaign has 24 hours and five available Slurm nodes. Inner runs
  have no separate time limit; the meta-evolver orchestrates their duration
  and scheduling. \\
\bottomrule
\end{tabular}
\end{table}

\paragraph{What happens in one concrete meta-evaluation.}
The outer evolver first edits \texttt{target/SKILL.md}, then invokes
\texttt{run.sh} with \texttt{--inner-env arc\_agi\_3}. The meta-environment
creates a fresh run containing the ARC-AGI-3 worker contract and installs the
candidate evolution skill in the inner evolver. That evolver edits
\texttt{agent/arc3\_agent/}, chooses games and repetitions, and evaluates
worker versions using Relative Human Action Efficiency (RHAE). The complete
inner run---worker versions, scores, traces, costs, and its nominated best
worker---returns as evidence to the outer evolver, which uses it to revise the
skill.

\paragraph{The key distinction.}
At the worker level, a candidate is an agent and an evaluation is a set of game
plays. At the meta-level, a candidate is an evolution policy and an evaluation
is an entire worker-evolution trajectory. The worker environment ranks matched
agents with a benchmark metric; the meta-evolver judges both the workers and
the quality and cost of the process that produced them. The interface is
shared, but the target, evaluation unit, and acceptance decision are not.

\section{Prompts}
\subsection{Canonical Seed Evolution Skill}
\label{app:seed-skill}

The evolver is stored as a versioned \texttt{SKILL.md} file. Every reported
meta-evolution lineage starts from the same canonical seed. The seed is
task-agnostic but not has instruction on how to interact with environement and be punctual with the time budget so we can control the experiments. The following excerpts preserve its
operative language while omitting repeated persistence reminders.

\begin{tcolorbox}[title=Canonical seed: interface and evidence (abridged)]
\ttfamily\small
Read \texttt{./ENVIRONMENT.md}. It states the goal, how to run an evaluation,
and where the results appear. It is the source of truth for what counts as
better. \dots{} The evaluation always runs the named agent: edit it in place,
and leave it where it is. \dots{}
\end{tcolorbox}

\begin{tcolorbox}[title=Canonical seed: search and handoff (abridged)]
\ttfamily\small
\dots{} Spend the
available wall-clock budget and reserve enough time to leave the work in order.
When time is up, leave the strongest version at \texttt{best/} and a short
\texttt{RESULT.md} stating what it is and what evidence says it is strongest.
\end{tcolorbox}

\begin{table}[H]
\caption{What the canonical seed fixes and what remains available for
meta-evolution to learn.}
\label{tab:seed-scope}
\begin{center}
\small
\setlength{\tabcolsep}{5pt}
\renewcommand{\arraystretch}{1.08}
\begin{tabular}{@{}p{0.46\linewidth}p{0.46\linewidth}@{}}
\toprule
Fixed by the canonical seed & Left open to the evolver \\
\midrule
Environment description as source of truth; edit only the declared target
& Which tasks or failure clusters to evaluate next \\
Preserve raw runs and artifacts; make claims proportional to evidence
& Evaluation configuration, replicate allocation, and numerical promotion tests \\
Use the available time; continue from the best-supported target
& Branching, parallelism, backtracking, composition, and search-layer changes \\
Return the strongest target under \texttt{best/} with an evidence summary
& Internal candidate organization and the path used to reach the nominee \\
\bottomrule
\end{tabular}
\end{center}
\end{table}

We observe that the evolver does not necessarily put the harness performing the best in the best folder. It has its own way to decide the delivered harness by including the total number of evaluation, paired comparison, etc, into consideration, showing that the evolver does more than maximizing benchmark scores.

\subsection{Meta-Evolved Instruction}

The meta-evolved version adds a more detailed experimental procedure. It
instructs the learner to establish a baseline, diagnose failed trajectories,
test different types of changes, account for noisy measurements, retain
candidate versions, and restore the best-supported target before finishing.

\begin{tcolorbox}[title=Meta-evolved instruction: evidence about the learner (abridged)]
\ttfamily\small
Treat an evaluation as evidence about the learner. Read the inner run, including
its raw measurements and trace, before deciding what to change in the evolver
instructions. Nominate the best-supported version; do not confuse the last
edited state with the best version.
\end{tcolorbox}

\end{document}